\documentclass[lettersize,journal]{IEEEtran}
\usepackage{amsmath,amsfonts}
\usepackage{algorithmic}
\usepackage{algorithm}
\usepackage{array}
\usepackage[caption=false,font=normalsize,labelfont=sf,textfont=sf]{subfig}
\usepackage{textcomp}
\usepackage{stfloats}
\usepackage{url}
\usepackage{verbatim}
\usepackage{graphicx}
\graphicspath{{./images/}}
\usepackage{booktabs,multirow}
\usepackage{cite}
\begin{document}

\title{Coordinate-Aware Thermal Infrared Tracking Via Natural Language Modeling}

\author{Miao Yan, Ping Zhang, Haofei Zhang, Ruqian Hao, Juanxiu Liu, Xiaoyang Wang, Lin Liu
        % <-this % stops a space
\thanks{This work was supported in part by the National Natural Science Foundation of China under Grant 62075031.}% <-this % stops a space
\thanks{Miao Yan, Ping Zhang, Ruqian Hao, Juanxiu Liu, and Lin Liu are with School of Optoelectronic Science and Engineering, University of Electronic Science and Technology of China, Chengdu 611731, China (e-mail: 2455442734@qq.com; pingzh@uestc.edu.cn; ruqianhao@uestc.edu.cn; juanxiul@uestc.edu.cn; liulin1979@uestc.edu.cn).}
\thanks{Ping Zhang is with the School of Optoelectronic Science and Engineering, Shenzhen Institute for Advanced Study, and the Yibin Institute, University of Electronic Science and Technology of China, Chengdu 611731, China (e-mail: pingzh@uestc.edu.cn).}
\thanks{Xiaoyang Wang is with the Innovation Centre, Office K1, Department of Computer Science, University of Exeter, Streatham Campus, Exeter, UK, EX4 4RN (e-mail: x.wang7@exeter.ac.uk).}
\thanks{Haofei Zhang is with NO.208 research institute of China Ordnance Industries, Beijing 102200, China(e-mail: jeamszhf@163.com).}
}

% The paper headers
\markboth{}%
{Shell \MakeLowercase{\textit{et al.}}: A Sample Article Using IEEEtran.cls for IEEE Journals}

\IEEEpubid{}
% Remember, if you use this you must call \IEEEpubidadjcol in the second
% column for its text to clear the IEEEpubid mark.
\maketitle

\begin{abstract}
Thermal infrared (TIR) tracking is pivotal in computer vision tasks due to its all-weather imaging capability. Traditional tracking methods predominantly rely on hand-crafted features, and while deep learning has introduced correlation filtering techniques, these are often constrained by rudimentary correlation operations. Furthermore, transformer-based approaches tend to overlook temporal and coordinate information, which is critical for TIR tracking that lacks texture and color information. In this paper, to address these issues, we apply natural language modeling to TIR tracking and  propose a coordinate-aware thermal infrared tracking model called NLMTrack, which enhances the utilization of coordinate and temporal information. NLMTrack applies an encoder that unifies feature extraction and feature fusion, which simplifies the TIR tracking pipeline. To address the challenge of low detail and low contrast in TIR images, on the one hand, we design a multi-level progressive fusion module that enhances the semantic representation and incorporates multi-scale features. On the other hand, the decoder combines the TIR features and the coordinate sequence features using a causal transformer to generate the target sequence step by step. Moreover, we explore an adaptive loss aimed at elevating tracking accuracy and a simple template update strategy to accommodate the target's appearance variations. Experiments show that NLMTrack achieves state-of-the-art performance on multiple benchmarks. The Code is publicly available at \url{https://github.com/ELOESZHANG/NLMTrack}.
\end{abstract}

\begin{IEEEkeywords}
Thermal infrared object tracking, Natural language model, Transformer tracking.
\end{IEEEkeywords}

\section{Introduction}
\IEEEPARstart{T}{HERMAL} infrared (TIR) object tracking is a fundamental and challenging task in artificial intelligence\cite{zhou2021object,pi2021instance}. Unlike visual imaging, TIR imaging captures the thermal information, and the TIR trackers can work effectively even in adverse conditions such as rain, fog, and low light. Therefore, TIR tracking has many applications in intelligent surveillance, maritime rescue, unmanned aerial vehicle platforms, etc. However, due to the low SNR and the lack of rich color information in TIR images, it is crucial to extract discriminative features and use temporal information as complementary information to the TIR features.

The original TIR trackers were based on traditional machine learning methods. For TIR images without color information, Jian and Jie et al. use a one-dimensional filter to filter the two dimensions of the TIR image separately, improving the performance of the tracker\cite{jian2004real}. Venkataraman et al. (2012) estimate the appearance representation of the target using pixel intensity histograms and the distribution of local standard deviations and verify that the model can improve tracking accuracy in multiple sequences\cite{venkataraman2012adaptive}. However, hand-crafted features lack adaptability, and the low-level feature representation is not sufficient to handle complex and unpredictable real-world scenarios.

 \IEEEpubidadjcol CNN and Transformer have demonstrated powerful representation capabilities in computer vision tasks and have been increasingly applied to TIR tracking. These TIR trackers can be generally classified into correlation filtering-based algorithms and Siamese network-based algorithms. A correlation filtering-based algorithm can be seen as a matching task. For instance, MCFTS\cite{liu2017deep} utilises different levels of CNN features and applies them to a correlation filtering framework to obtain multiple response maps fused by KL divergence to enhance tracking robustness. MLSSNet\cite{liu2020learning} can simultaneously learn global semantic and local detail information to improve discriminative ability. However, these methods only use convolutional neural networks to extract features, which proves suboptimal for infrared imagery characterized by insufficient texture detail. And they are challenging to track robustly when facing the target being occluded or reappearing after disappearing. Inspired by the visual tracking method, the Siamese network-based algorithm is gradually adopted for TIR object tracking, which can be viewed as a detection task. SiamMSS\cite{li2022multigroup}  introduces a spatial shift model under the Siamese network framework to boost the feature expression ability of the network and uses split attention to fuse the feature maps. To learn diverse TIR features, Yang et al.\cite{yang2024learning} further enhanced the Transformer network and designed a diversity loss function and an auxiliary branch. Despite the outstanding performance of these models, they are complex in design and neglect the temporal correlation.
 
In recent years, some visual tasks have been redefined as sequence generation tasks through the direct application of natural language models. This paradigm shift has introduced a natural language interface to the realm of visual tasks. A pioneering example of this approach is Pix2Seq\cite{chen2021pix2seq}, which transforms target detection into a coordinate-generative task—a first in the field.  Building on this innovation, Seqtrack\cite{chen2023seqtrack} leverages an autoregressive sequence-to-sequence model to generate target sequences and achieve optimal performance in visual tracking. Drawing inspiration from these advancements, to overcome the lack of texture detail information in infrared video sequences, we propose a novel framework that integrates coordinate information and temporal correlation as auxiliary information alongside infrared features, thereby augmenting the robustness and accuracy of TIR tracking.

In this paper, different from the previous TIR object tracking model, we formulate TIR object tracking as a language modeling task based on coordinate sequence generation, thereby making more effective use of coordinate and time information. We construct an encoder-decoder transformer language model that abandons complex head networks. It simplifies the tracking pipeline and leverages both time series and coordinate information as auxiliary information for TIR features to enhance tracking robustness. Specifically, the encoder unifies feature learning and relational modeling, while the decoder predicts bounding box values step by step by combining coordinate information with the TIR features. We present a multi-level progressive fusion module to enhance semantic information and introduce multi-scale features to tackle challenges such as background interference and target scale variations. We also explore a simple template update strategy to assess the reliability of the dynamic template and design an adaptive loss function to enhance tracking performance. We evaluate NLMTrack on several datasets, including the PTB-TIR\cite{liu2019ptb} dataset, the VOT-TIR2015\cite{felsberg2015thermal} dataset, and the LSOTB-TIR\cite{liu2023lsotb} dataset, and demonstrate its superior performance over existing methods. The key contributions of our work are outlined as follows:

\begin{itemize}
\item We introduce natural language models into the TIR object tracking framework, which is the first time that TIR object tracking is cast as a language modeling task based on coordinate sequence generation.
\item We devise a multi-level progressive fusion module to enrich the semantic representation and introduce multi-scale features. The module produces features at different levels through a simple feature pyramid and gradually fuses cross-semantic features using our fusion modules UpFusion and DownFusion. This method enhances the semantic understanding of the target and preserves fine-grained information at different scales.
\item Our model training aims to maximize the log-likelihood of the target sequence, based on which we integrate the SIOU loss. Meanwhile, we only use the mean of the softmax scores of the predicted target sequences to assess the reliability of the dynamic template, without requiring two-stage training.
\item Our proposed tracking model performs best on multiple challenging benchmarks compared to other TIR tracking algorithms.
\end{itemize}

\section{Related Work}

\subsection{TIR Tracking}
TIR imaging technology has developed rapidly in recent years, and many researchers have made significant progress in the growing interest in TIR object tracking. TIR object tracking methods can be generally divided into two groups: machine learning-based methods, which rely on hand-crafted features, and the deep learning-based methods, which learns features from data. Gao et al.\cite{gao2016infrared} combine three features, including local entropy, local contrast mean difference, and intensity histogram, to achieve a multi-feature joint sparse representation for TIR object tracking. ABCD\cite{berg2016channel} is a tracking method based on an enhanced distributed field representation. It uses channel coding features to enhance the distributed field representation and does not rely on color features and target contour information, making it suitable for TIR object tracking. However, ABCD fails to handle challenges such as distortion of the target and occlusion. Meng et al.\cite{ding2019thermal} present a method for TIR object tracking that fused directional gradient histograms and normalized grayscale features to calculate response maps. TIR-MS\cite{yun2019tir} enhances the robustness of TIR object tracking by incorporating temperature data as auxiliary information along with the image luminance features. CODIFF\cite{demir2016co} reduces the computational complexity while preserving the benefits of the covariance matrix, and it naturally integrates the information from the histogram features and the epistatic model. However, due to the low-level feature representation of hand-crafted features and the lack of adaptability, traditional machine learning-based methods are challenging to deal with complex and unpredictable real-world scenarios.

CNN and Transformer\cite{liu2022learning} are more powerful in feature representation than hand-crafted features. Deep learning-based methods for TIR object tracking can be classified into correlation filter (CF)-based methods and Siamese network-based methods. The algorithm predicated on correlation filtering is conceptualized as a task of similarity match. Within this framework, a CNN is utilized to delineate the features from both the search region and the initial template. Subsequently, these extracted features are assimilated into the CF framework. This integration facilitates the computation of a response map, wherein the peak value signifies the precise location of the target.  ECO-TIR\cite{zhang2018synthetic} generates synthetic TIR video sequences for neural network training under the ECO\cite{danelljan2017eco} framework, which improves the network representation capability and consequently enhances the performance of TIR object tracking. HSSNet\cite{li2019hierarchical} integrates the shallow features with the deeper semantic features and applies the attention mechanism to boost the representation and discrimination ability of the network. Furthermore, multi-scale inputs are adopted to handle the problem of target scale variations. Ding et al.\cite{ding2022thermal} fuse hand-crafted and convolutional features through a novel search response map strategy to fully and effectively exploit the complementary information of the two. Parhizkar et al.\cite{parhizkar2023object} employ a local directional number pattern algorithm to overcome the interference of the target by the complex environment and introduce a target attention to make the model focus on important information. MMNet\cite{liu2022learning}, a two-layer feature model based on global attention and convolutional neural networks, constructs a novel framework for TIR object tracking by learning discriminative features in TIR images and fine-grained features at the pixel level. AMFT\cite{yuan2023robust} integrates color name (CN), HOG, and CNN features under the framework of correlation filtering. This method exploits the complementary information and discriminative information of multiple features and thus improves the feature representation capability. However, online algorithms based on correlation filtering inherently limit the richness of models they can learn\cite{bertinetto2016fully}.

The Siamese network-based algorithm can be regarded as a detection task to achieve accurate target localization and tracking through feature extraction, fusion, and detection. GFSNet\cite{chen2022gfsnet} proposes a category adaptive module and a localization adaptive module, and the adaptive module is verified to have strong characterization and generalization capabilities on multiple datasets. SiamMSS\cite{li2022multigroup} introduces a spatial offset model under the framework of Siamese networks, which enhances the detailed information of feature maps by grouping and offsetting the feature maps and fusing features using mutual correlation operation. However, it is difficult to fully express the similar semantic information between the two by fusing template features and search region features using simple correlation operations. Inspired by Transformer, Zhao et al.\cite{zhao2022thermal} utilize Transformer with global modeling capability for feature extraction and fusion, which improves the model’s ability to attend to contextual information and focus on useful semantic information. Meanwhile, to fully utilise the ability of Transformer to expand the training set with the generated TIR sequences. Yang et al.\cite{yang2024learning} propose a TIR object tracking model that can learn diverse TIR features, designing a two-branch structure to capture the fine-grained features while presenting an adaptive loss function to reinforce the learning of fine-grained features. Although these Siamese-network-based algorithms achieve good performance, they usually neglect the temporal information and the coordinate information within the frame, which are particularly important for TIR object tracking that lacks texture information and color.

\subsection{Natural Language Models}
Natural language modeling has witnessed significant advances in recent years, with the emergence of the Generative Pre-trained Transformer (GPT) language model causing a paradigm shift in several domains\cite{brown2020language}. Various generative models that integrate natural language and computer vision have been developed for tasks that involve multimodal data. Videobert\cite{sun2019videobert} is a joint model that learns language and video representations based on the BERT model. It is capable of modeling the joint distributions of the two modalities, enabling several tasks such as text-to-video generation and future frame prediction. CogView2\cite{ding2022cogview2} is a hierarchical transformer that also adopts parallel autoregression to enhance the efficiency and quality of the generated images. Recently, natural language models have been explored for direct application to vision tasks. Pix2Seq\cite{chen2021pix2seq} solves the target detection task through language modeling. The target is “described” by an encoder, and the target sequences are generated by a decoder that fuses the visual features of the encoder and the previous tokens. The algorithm offers a natural language interface for computer vision tasks. Seqtrack\cite{chen2023seqtrack} is a method that uses a sequence-to-sequence model to predict bounding boxes in an autoregressive way, and the loss employs only a cross-entropy loss. Inspired by Seqtrack, we present a sequence generative model that directly predicts the bounding box. Our method decodes the object coordinates progressively through intra-frame prediction, dispensing with the complex bounding box regression branch and classification branch. Meanwhile,  the encoder integrates feature extraction and fusion. This simplifies the tracking framework and presents a novel solution idea for TIR object tracking.

Our coordinate sequence generative framework shares similar ideas with SeqTrack. They both convert a tracking task into a coordinate sequence generation task. However, our approach is distinct from SeqTrack in four fundamental aspects. 1) The encoder structure is different. SeqTrack uses ViT\cite{chen2023seqtrack} as the encoder for feature extraction, which is computationally expensive. Our encoder prioritises encoders with intrinsic features of the template, thus mitigating the impact of redundant interactions on the integrity of the template. In addition, the encoder's emphasis on template-intrinsic features helps to improve the interaction of target information within the search region, thereby improving efficiency and effectiveness. 2) To enrich the semantic information and integrate features across multiple scales, we develop a multi-level progressive fusion module. Employing a hierarchical fusion approach ensures a comprehensive synthesis of information, thereby enhancing the model's ability to interpret and process complex TIR scenarios.3) The loss functions are different. SeqTrack is designed for visual images, while ours is tailored for TIR images, which have different characteristics and challenges. We introduce the SIOU\cite{gevorgyan2022siou} loss, which constrains the spatial properties of the bounding box, to enhance the performance of TIR tracking.

\section{Proposed Method}
This section delineates the overall architecture and integral components of our proposed model. Firstly, we briefly overview our natural language modeling-based TIR object tracking framework. Then, we detail the encoder, the multilevel progressive fusion module, and the decoder. Lastly, we introduce the design of the adaptive loss during training and the dynamic template update scheme.

\begin{figure*}
    \centering % 表示居中
    \includegraphics[width=18cm,height=9cm]{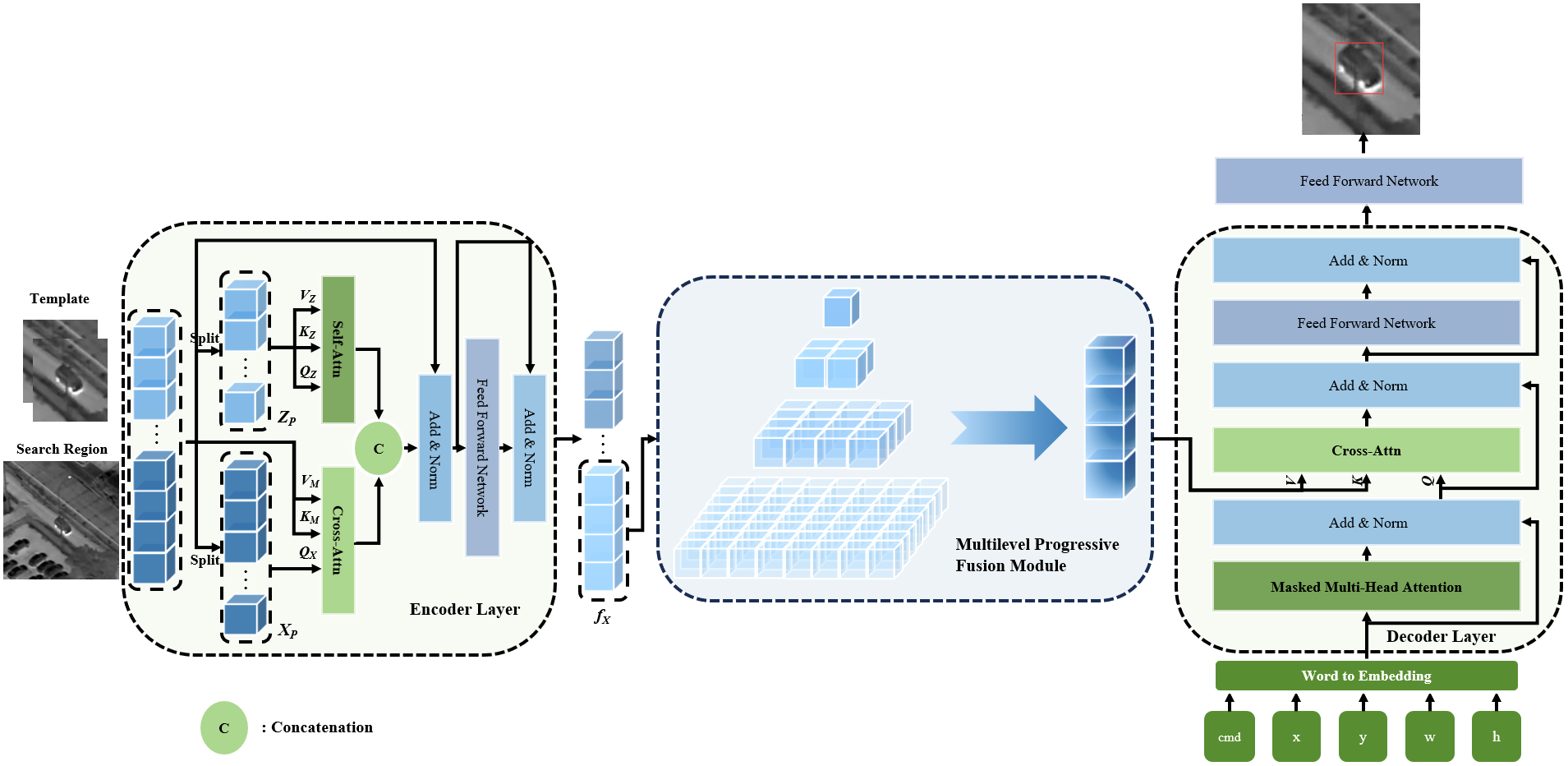}
    \caption{Our NLMTrack framework. The overall framework consists of an encoder, a multilevel progressive fusion module, and a decoder with a causal Transformer. $ Z_p$ and $ X_p$ refer to TIR features extracted from the template and the search region, respectively. $ f_x$ denotes the search region features from the encoder’s output.} \label{overview}
\end{figure*}

\subsection{Algorithm Overview}
We present a natural language modeling-based TIR object tracking model named NLMTrack, whose architecture is illustrated in Fig. \ref{overview}. NLMTrack mainly comprises an encoder, a multilevel progressive fusion module, and a decoder. Specifically, the encoder we adopt can concurrently conduct feature extraction and feature fusion. After that, the fused features are passed to a multilevel progressive fusion module to acquire more expressive semantic features while combining multi-scale features. Finally, the bounding box values are incrementally generated by fusing the TIR features and coordinate sequence information using the multi-attention module with a mask and the cross-attention module. In response to the challenges posed by the target's appearance variations, our research introduces a straightforward yet effective template reliability assessment strategy.

\subsection{Encoder for Unified Feature Extraction and Feature Fusion}
We employ the Transformer-based encoder-decoder, which is prevalent in vision tasks and language modeling \cite{brown2020language}, where the encoder is primarily responsible for feature learning of TIR images. Siamese network-based methods for TIR tracking employ two separate modules to extract and fuse features in a sequential manner, which restricts the information exchange, leading to poor discriminative ability between the object features and the background. To avoid this issue, we adopt a Transformer structure to extract features while achieving the fusion of $ Z_p$ and $ X_p$, as shown in the encoder part in Fig. \ref{overview}.

The encoder of NLMTrack takes the template $ I_z$ and the search image $ I_x$ as inputs. $ I_z$ is obtained by cropping and resizing a region from the original image, centered on the previously tracked bounding box, and scaled by a factor of four. The final size of  $ I_z$ is $ (H_z, W_z)$. Similarly, the search region $ I_x$ is cropped and resized from the current frame to be predicted, with an area $ 4.5^2$ times the object bounding box size, to $ (H_x, W_x)$. Following the Visual Transformer \cite{yuan2021temporal,wei2023autoregressive,cui2022mixformer}, we employ a convolutional embedding layer to map the two images into block embeddings, respectively, which are then flattened into one-dimensional patch sequences, $ z_p\in R^{N_z\times C}$and $ x_p\in R^{N_x\times C}$, where $ N_z=H_zW_z/16^2$ and $ N_x=H_xW_x/16^2$ refer to the number of $ z_p$ and $ x_p$, $ Z_p$ and $ X_p$ represent TIR features of $ z_p$ and $ x_p$. To incorporate positional information, we embed sinusoidal positional encodings separately for $ z_p$ and $ x_p$. And the two patch sequences are concatenated and fed into the encoder layers. Specifically, we first project $ Z_p$ and $ X_p$ into queries, keys, and values by linear mapping. We denote targets by $ Q_z$, $ K_z$, and $ V_z$, and the search regions by $ Q_x$, $ K_x$, and $ V_x$. $ Attn(\cdot)$ can be formulated as follows:
\begin{equation}
\begin{aligned}
& K_m=Concat(K_z, K_z), \quad V_m=Concat(V_z, V_z), \\
& {Attn}_{\mathrm{x}}=Softmax(\frac{Q_x K_m^T}{\sqrt{d}}) V_m, \\
& {Attn}_{\mathrm{z}}=Softmax(\frac{Q_z Q_z^T}{\sqrt{d}}) V_z, \\
& {Attn(\cdot)}={Concat}\left(Attn_z, Attn_z\right),
\end{aligned}
\end{equation}
where $ Attn_z$ is the self-attention result,$ Attn_x$ is the cross-attention result, and $ Attn$ is a feature sequence formed by concatenating $ Attn_z$ and $ Attn_x$. Then the concatenated feature is processed by LayerNorm layer and multilayer perceptron as:
\begin{equation}\label{eq1}
\begin{aligned}
& A^{i-1}=Attn(Norm(E^{i-1}))+E^{i-1}, \\
& E^i=MLP(Norm(A^{i-1}))+A^{i-1},
\end{aligned}
\end{equation}
where $ MLP( \cdot )$ denotes multilayer perceptron, $ Norm(\cdot)$ represents LayerNorm layer, $ E^i$ denotes the result of the $ i$-th encoder layer, $ E^0$ is a sequence formed by concatenating $ z_p$ and $ x_p$, $ E$ is the concatenation of $ Z_p$ and $ X_p$ , and $ Attn(\cdot)$  refers to the Transformer-based attention mechanism. By employing self-attention, we can model the appearance of the template more comprehensively and utilize cross-attention to facilitate communication between the search region and the template. This strategy integrates the feature extraction and fusion processes. It prevents the need for the information aggregation module, thus attaining a high degree of parallelism and enabling the network to learn more discriminative features.

\subsection{Multilevel Progressive Fusion Module}
\begin{figure}
    \centering % 表示居中
    \includegraphics[width=8cm,height=4cm]{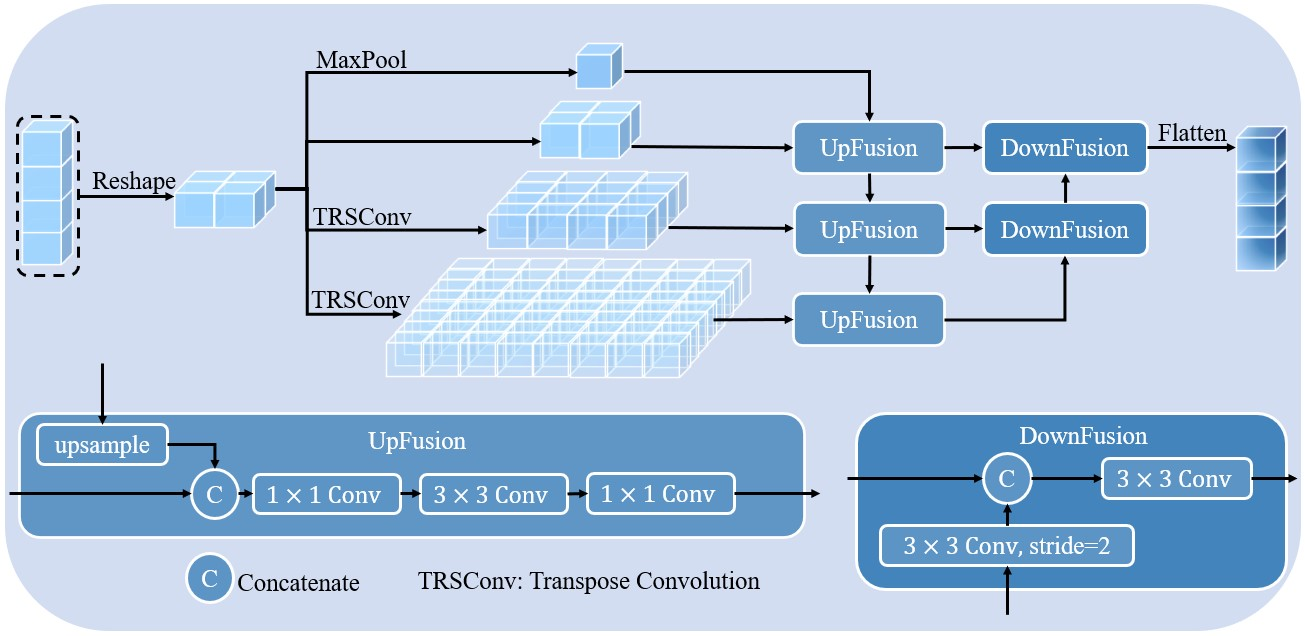}
    \caption{Detailed architecture of the multilevel progressive fusion module. The module generates features at different levels through a simple feature pyramid and progressively fuses cross-semantic features using our fusion modules UpFusion and DownFusion.} \label{MPFM}
\end{figure}

TIR object tracking often encounters challenges such as background interference and target scale variations, while rich semantic information assists in better discriminating between target and background, and multi-scale features aid in adapting to scale variations. To tackle this challenge, we design a multilevel progressive fusion module. The encoder structure employed in this paper is a variant of Visual Transformer (ViT), where each layer of inputs and outputs has the same size and thus cannot naturally generate semantic information at different levels. Based on He et al.\cite{li2022exploring}, a self-attentive mechanism with global modeling capability under supervised training can learn scale-equivariant features, and thus pyramidal features can be obtained by only performing MaxPool operation and transposed convolution on the last layer of the feature map of the ViT. Inspired by this, as depicted in Fig. \ref{MPFM}, we first apply the same operation to acquire semantic features at different levels. Subsequently, we integrate the high-level semantic information with the high-resolution feature map at the lower end via the UpFusion module in a top-down manner. This is followed by a bottom-up transfer of the high-resolution feature details at the lower end to the semantic feature map at the upper end through the DownFusion module. The input and output sizes of the multilevel progressive fusion module are consistent to ensure the acquisition of different levels of multi-scale semantic information while avoiding the excessively large feature maps that impose too much computational burden on the subsequent decoder.

We propose the Upfusion module, which can be regarded as a gradual upsampling process, as depicted in Fig. \ref{MPFM}. The process for UPfusion is as follows:
\begin{equation}
\begin{aligned}
& F_{cat}=Concat(F_h,upsample(F_l), \\
& F_{up}=Conv_{1\times 1}(Conv_{3\times3}(Conv_{1\times1}(F_{cat})))),
\end{aligned}
\end{equation}
where the UPfusion module combines the deep and shallow feature maps, denoted by $ F_h$ and $ F_l$, respectively. To match the spatial resolution of $ F_h$, $ F_l$ is up-sampled by a factor of two using bilinear interpolation. Then, $ F_h$ and $ F_l$  are concatenated, resulting in $ F_{cat}$. $ Conv_{k\times k}$ refers to a $ k\times k$ convolution layer. From this, the UPfusion module can utilize the information from the context features to generate novel multi-scale features. In the Downfusion module, we use similar operations to achieve the fused features.

\begin{figure*}[!t]
\centering
    \subfloat[ConF]{
        \includegraphics[width=4.5cm,height=2.5cm]{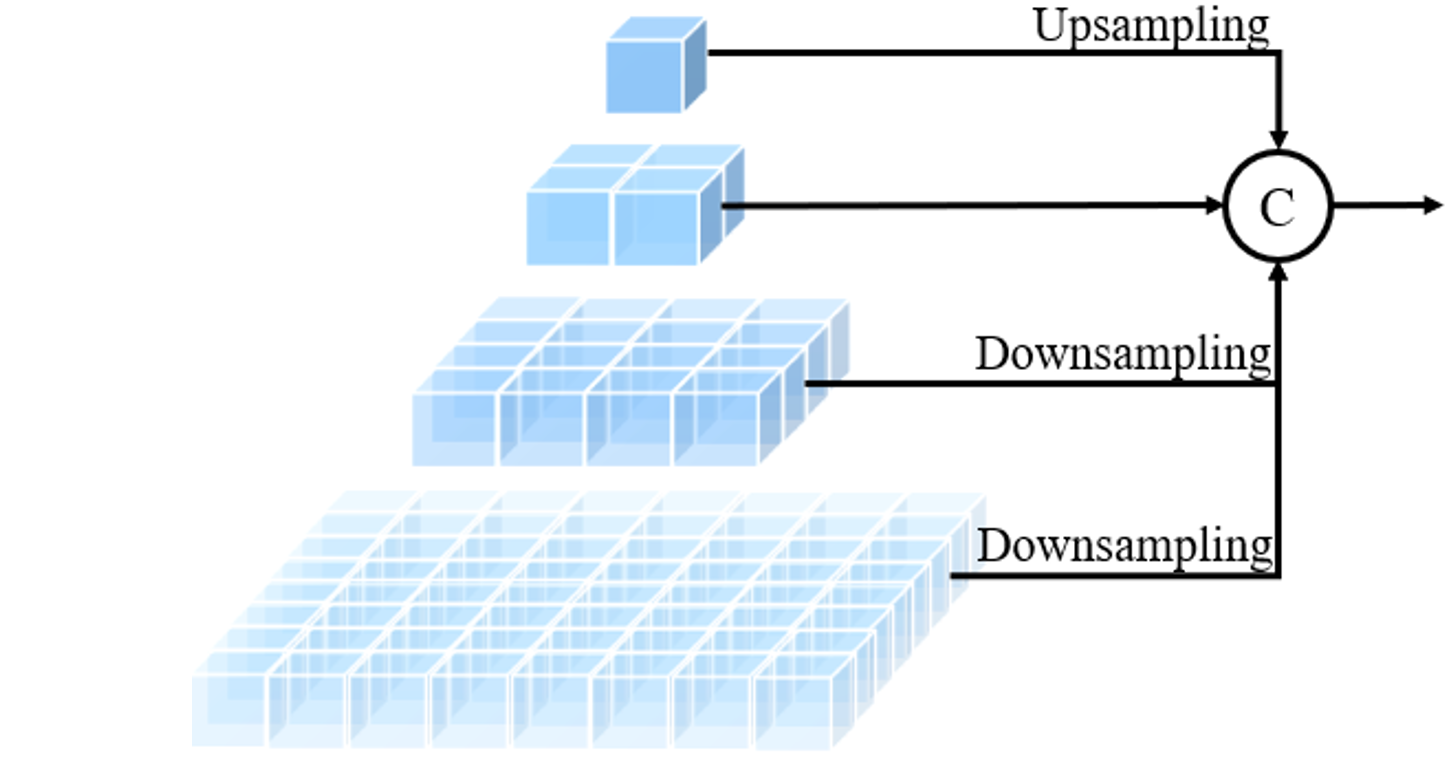} %
        \label{fusion(a)}} 
    \hfil
    \subfloat[AddF]{
        \includegraphics[width=4cm,height=2.5cm]{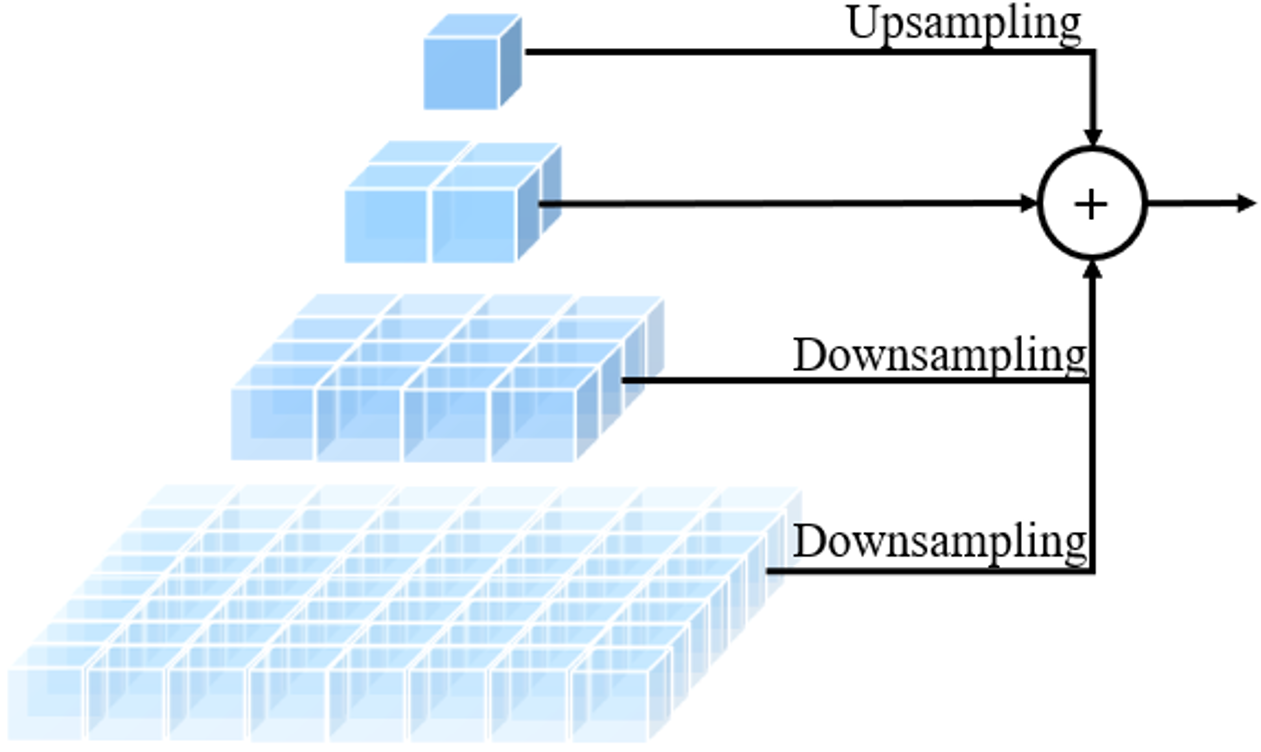} %
        \label{fusion(b)}}
    \hfil
    \subfloat[Ours]{
        \includegraphics[width=6cm,height=2.5cm]{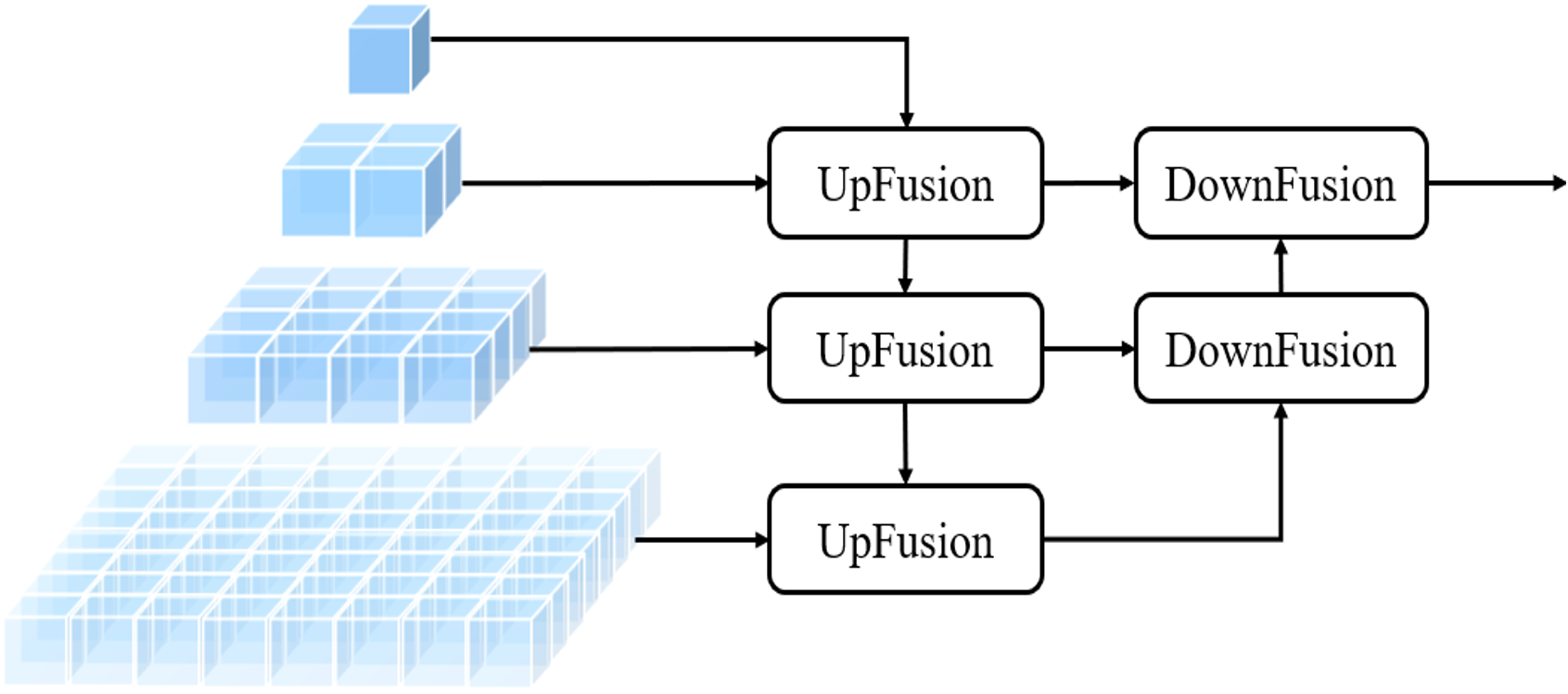} %
        \label{fusion(c)}}   
    \caption{Illustration of the structure of commonly used fusion methods with our proposed method. (a) Fusion in the form of concatenation; (b) Fusion in the form of element-wise addition operation; (c) Progressive fusion approach. } 
    \label{fusion}
\end {figure*}

To substantiate the efficacy of our novel progressive fusion methodology, we conduct a comparative analysis against two prevalent fusion methodologies, as seen in Fig. 3. The first fusion method entails resizing the feature maps of each layer in the pyramid feature to the same size as the second layer feature map (from deep to shallow) by upsampling or downsampling, and then conducting the concatenation and 1×1 convolution operations. The second fusion approach is analogous to the first, with the main difference being that the concatenation operation is substituted with an element-wise addition operation. Both upsampling and downsampling operations are identical to those in Fig. \ref{fusion}\subref{fusion(c)}. We exhibit the superiority of the progressive fusion approach in subsequent experiments.

\subsection{Decoder with a Causal Transformer}
The input to the decoder consists of a sequence of TIR features and a sequence of coordinates. This coordinate sequence is $ [cmd, x, y, w, h]$, where $ cmd$ corresponds to the learnable embedding specified in a shared vocabulary $ V$ for predicting the target sequence. We represent the bounding box by $ [x, y, w, h]$,  which is a common and intuitive way of describing the target. The coordinates need to be discretized into a sequence of tokens, specifically each value in $ [x, y, w, h]$ is quantized as an integer between $ [1, nbins]$. Each integer between $ [1, nbins]$ corresponds to a learnable embedding in the shared vocabulary $ V$. The shared vocabulary is iteratively optimized during training.

Our architecture incorporates a masked multi-head attention module, as visualized in the right part of Fig. \ref{overview}. This module utilizes a self-attention mechanism equipped with a causal mask, guaranteeing that the output for each element in the sequence is influenced exclusively by its preceding tokens. Specifically, the masked self-attention mechanism first calculates the similarity between the query and the fundamental matrices. Then it adds an upper-triangular Mask matrix, where the non-zero values are negative infinity, which enables the use of the sequence information before the current token and prevents focusing on the sequence information after it. Next, the coordinate features and TIR features are fused using the cross-attention module. Moreover, the fused features are further augmented by a feed-forward network, and the augmented features are used as inputs to the next decoder layer.

At inference time, The target sequence generation begins with the start token $ cmd$ as the input to the decoder. The decoder then samples one by one from the shared vocabulary to obtain a target sequence token of the form $ [x, y, w, h, end]$. This prediction offset and a causal mask ensure that NLMTrack is autoregressive. Since only the coordinates are needed in our task, there is no need to use the additional terminator $ end$ to terminate the generation process during inference. Finally, we de-quantize the discrete tokens of the generated target sequence to obtain continuous coordinate values. The advantage of this step-by-step approach to decoding the target bounding box is that it relies on TIR features and also leverages previous coordinate sequences as prompt information, leading to more accurate TIR object tracking. Meanwhile, it simplifies the model structure by dispensing with customized classification headers and bounding box regression branches.

\subsection{Loss and Dynamic Template Update Scheme}
\subsubsection{Loss}NLMTrack treats TIR object tracking as a coordinate sequence generation-based language modeling task, which can be expressed as:
\begin{equation}
P\left({B}^t \mid {B}^{0: t-1},\left({C}, {z}^t, {x}^t\right)\right),
\end{equation}
where $ C$ is the start token $ cmd$. $ z^t$ is the template image at the current moment $ t$, which is dynamically updated over time. $ x^t$ denotes the image of the search region corresponding to the current moment $ t$, $ {B}^{0: t-1}$ is the target sequence previously predicted at the current moment $ t$ and is related to $ z^t$ and $ x^t$.

Similar to the objective function of the natural language modeling task, we maximize the log-likelihood of the generated sequences through the softmax cross-entropy loss function:
\begin{equation}
L_{C E}={Max} \sum_{t=1}^T \log P\left(B^t \mid B^{0: t-1},\left(z^t, x^t\right)\right) \text {, }
\end{equation}
where $ T$ is the length of the target sequence, $ t$ corresponds to the time step of the token, and $ P(\cdot)$ represents the softmax probability. $ Max$ denotes maximize the log-likelihood.

The cross-entropy loss is limited by its neglect of the spatial attributes inherent to the target. We investigated how to integrate the coordinates information into the cross-entropy loss paradigm to enhance the tracking performance. To this end, we have integrated the SIOU loss, as proposed by Gevorgyan et al.\cite{gevorgyan2022siou}, which meticulously accounts for discrepancies in area, shape, and orientation between the predicted bounding boxes and the ground truth. This inclusion enriches the loss function with spatial awareness, thereby refining the tracking performance. The comprehensive loss function, encapsulating these considerations, is expressed as follows:
\begin{equation}
L=L_{CE}+L_{SIOU}
\end{equation}
where $ L_{SIOU}$ denotes the SIOU loss function. Since sampling from a shared vocabulary cannot be back-propagated during training, we use the maximum of the softmax scores of the generated marker sequences to represent the coordinate information. Although it is independent of the task objective, training in conjunction with this prior knowledge proved effective in subsequent experiments. 

\subsubsection{Dynamic template update scheme}To handle target occlusion or deformation during tracking, we introduce a dynamic template updating strategy. The templates are first extracted from the first frame and kept constant. Then, another dynamically updated template is added to adapt to the target's appearance variations. This scheme not only avoids updating a large number of parameters over time but also leverages temporal information to improve the stability of TIR object tracking, which is essential for TIR tracking.

The quality of the templates affects the tracking performance significantly\cite{yan2021learning}. Hence, we devise a score evaluator to assess the reliability of the templates, which evaluates the targets based on the probability of the generated tokens. We only need to compute the mean of the softmax scores of the predicted coordinate values $ [x, y, w, h]$ to select reliable online update templates. The dynamic template is initialized with the fixed template of the starting frame. It is updated only when the average softmax score exceeds the threshold $ \lambda$ while the specified number of frames between updates $ Z_u$ is reached. Otherwise, it remains unchanged. $ \lambda$ is set to 0.6, which can be tuned based on the difficulty of the tracked video sequences, and the $ Z_u$ interval frames to 25. Unlike previous work\cite{yan2021learning,cui2022mixformer}, NLMTrack eliminates the need to construct an additional score prediction branch and avoids the complexity of two-stage training by simply computing the mean value to assess the reliability of the template.

\section{Experimental Results}
In this section, we first present the implementation details,  the datasets, and the metrics employed for evaluation. Furthermore, we compare the NLMTrack tracker with other state-of-the-art trackers and demonstrate NLMTrack's superior performance. We furnish a selection of visual results, which serve to exemplify the NLMTrack tracker's proficiency in addressing complex tracking challenges with efficacy. Lastly, ablation studies are conducted to validate the individual contributions of the various components within our model.

\subsection{Implementation Details}

\subsubsection{Model}The encoder is identical to ViT with 12 layers, each patch size is 16$ \times$16, and the decoder comprises the same two decoder layers. We set the quantized nbins and V size to 4000. The learnable embedding dimension in the shared vocabulary is set to 256, the same as the hidden layer dimension of the decoder. We used a three-layer perceptron to sample words from the embedding layer, which also has a hidden layer dimension of 256 and an output layer dimension of nbins, the same size as the shared vocabulary table V. The softmax layer is used to select the maximum likelihood words as output.

\subsubsection{Training}The training data consists of LSOT-TIR\cite{liu2023lsotb} and the visible dataset GOT-10k\cite{huang2019got}. Due to the addition of a dynamically updated template, we sample two templates and a search region image as a set. We set $ (H_z, W_z)$ to (128, 128) and $ (H_x, W_x)$ to (288, 288). The model is trained using the AdamW optimizer, setting the learning rate to $1 e^{-5}$ for the encoder, and $1 e^{-4}$ for the non-encoder module. We train 120 epochs, sampling 30,000 sample pairs per epoch with a batch size of 16. The proposed NLMTrack tracker is trained and evaluated using an NVIDIA GeForce RTX 3090 GPU, achieving a speed of 33 FPS.

\subsection{Datasets and Evaluation Metrics}
The evaluation datasets we used include VOT-TIR2015\cite{felsberg2015thermal}, PTB-TIR\cite{liu2019ptb}, and LSOTB-TIR\cite{liu2023lsotb} datasets. The VOT-TIR2015 benchmark is the first dataset that provides a platform of codes for performance evaluation. It comprises 20 sequences with labeling information, containing 11,000 images, covering six challenge attributes such as occlusion, camera motion, size transformation, etc. Its three the metrics employed for evaluation are Robustness (Rob),  Accuracy (Acc), and Expected Average Overlap (EAO). Rob measures the stability of the tracker by calculating the average frequency of tracking failures. EAO is a composite metric, which allows for the joint assessment of the accuracy and stability of the tracker. The PTB-TIR dataset covers almost all the pedestrian tracking scenarios in the civil sector, such as surveillance, UAV-borne, handheld, and vehicle-borne scenarios, and consists of 60 pedestrian video sequences, over 30,000 images, and nine different challenging attributes. The PTB-TIR benchmark evaluates the tracker using two main metrics: success rate (Suc) and precision rate (Pre). The LSOTB-TIR dataset consists of 1,280 video sequences with four scenarios, containing 12 challenging attributes such as thermal crossover, camera shake, and target intensity change. In addition to the PTB-TIR evaluation metrics,  LSOTB-TIR also employs the normalized precision rate (NormP).

\subsection{Evaluation on VOT-TIR2015 Dataset}
To prove the superiority of NLMTrack, we evaluate it with 13 advanced trackers on the VOT-TIR2015 dataset. These include commonly used trackers such as TBOOST\cite{gundogdu2015comparison}, SRDCF\cite{danelljan2015learning}, Siamese-FC\cite{bertinetto2016fully}, CFNet\cite{valmadre2017end}, MCFTS\cite{liu2017deep}, ECO-deep\cite{danelljan2017eco}, HSSNet\cite{li2019hierarchical}, MLSSNet\cite{liu2020learning}, ECO-MM\cite{liu2022learning}, MMNet\cite{liu2022learning}, as well as the latest TIR object tracking algorithms proposed in the last two years, such as GFSNet\cite{chen2022gfsnet}, SiamMSS\cite{li2022multigroup}, and DFG\cite{yang2024learning}.

Table \ref{vot} presents that our NLMTrack outperforms the other trackers, reaching the best EAO of 0.392, Acc of 0.78, and Rob of 2.02. Compared to SiamMSS\cite{li2022multigroup}, NLMTrack improves by 1.5\% in EAO score. Notably, our tracker outperforms the tracker MMNet\cite{liu2022learning} and GFSNet\cite{chen2022gfsnet} with absolute gains of 0.8\% in Robustness scores. While NLMTrack achieves the same results as DFG\cite{yang2024learning} in Accuracy scores, our advantage in EAO scores and Robustness scores is very significant, suggesting that our tracker has a higher level of robustness.
\begin{table}[!t]
  \centering
  \caption{Evaluation results of NLMTrack and 13 other trackers on VOT-TIR2015 benchmark. The up arrow denotes a larger value is better performance. The best results are highlighted in bold.}
    \begin{tabular}{lllll}
    \toprule[0.6pt] 
\cmidrule{1-5}    \multicolumn{1}{p{8em}|}{Category} & \multicolumn{1}{p{5.625em}}{Tracker} & {EAO↑} &{Acc↑} & {Rob↓}\\
\cmidrule{1-5}    \multicolumn{1}{l|}{\multirow{2}[2]{*}{\begin{tabular}[c]{@{}l@{}}Traditional method \\ based trackers\end{tabular}}} & \multicolumn{1}{p{6.625em}}{TBOOST\cite{gundogdu2015comparison}} & 0.192 & 0.56  & 3.30 \\
    \multicolumn{1}{l|}{} & \multicolumn{1}{p{5.625em}}{SRDCF\cite{danelljan2015learning}} & 0.225 & 0.62  & 3.06 \\
\cmidrule{1-5}    \multicolumn{1}{l|}{\multirow{5}[1]{*}{\begin{tabular}[c]{@{}l@{}}Deep learning based \\ CF trackers\end{tabular}}} & \multicolumn{1}{p{5.625em}}{MCFTS\cite{liu2017deep}} & 0.218 & 0.59  & 4.12\\
    \multicolumn{1}{l|}{} & \multicolumn{1}{p{5.625em}}{CFNet\cite{valmadre2017end}} & 0.282 & 0.55  & 2.82\\
    \multicolumn{1}{l|}{} & \multicolumn{1}{p{5.625em}}{HSSNet\cite{li2019hierarchical}} & 0.311 & 0.67  & 2.53 \\
    \multicolumn{1}{l|}{} & \multicolumn{1}{p{5.625em}}{MLSSNet\cite{liu2020learning} } & 0.329 & 0.57  & 2.42  \\
    \multicolumn{1}{l|}{} & \multicolumn{1}{p{6.5em}}{ECO-deep\cite{danelljan2017eco}} & 0.286 & 0.64  & 2.36  \\
    \multicolumn{1}{l|}{} & \multicolumn{1}{p{6.5em}}{ECO-MM\cite{liu2022learning}} & 0.303 & 0.64  & 2.44 \\
    \multicolumn{1}{l|}{} & \multicolumn{1}{p{5.625em}}{MMNet\cite{liu2022learning} } & 0.344 & 0.61  & 2.10  \\
\cmidrule{1-5}    \multicolumn{1}{l|}{\multirow{3}[2]{*}{\begin{tabular}[c]{@{}l@{}}Siamese network-based \\ deep trackers\end{tabular}}} & \multicolumn{1}{p{6.8em}}{Siamese-FC\cite{bertinetto2016fully}} & 0.219 & 0.60   & 4.10 \\
    \multicolumn{1}{l|}{} & \multicolumn{1}{p{5.625em}}{GFSNet\cite{chen2022gfsnet}} & 0.365 & 0.67  & 2.10   \\
    \multicolumn{1}{l|}{} & \multicolumn{1}{p{5.625em}}{SiamMSS\cite{li2022multigroup}} & 0.377 & 0.73  & 2.75\\
\cmidrule{1-5}    \multicolumn{1}{l|}{\multirow{2}[2]{*}{Transformer based trackers}} & \multicolumn{1}{p{5.625em}}{DFG\cite{yang2024learning}} & 0.329 & \textbf{0.78} & 2.41  \\
    \multicolumn{1}{l|}{} & \multicolumn{1}{p{5.625em}}{NLMTrack(ours)} & \textbf{0.392} & \textbf{0.78} & \textbf{2.02} \\
       \bottomrule[0.7pt] 
    \end{tabular}%
  \label{vot}%
\end{table}%

\subsection{Evaluation on PTB-TIR Dataset}
\subsubsection{Overall Performance}Table \ref{ptb} presents the superiority of NLMTrack on the PTB-TIR dataset. These compared trackers comprise the widely used KCF\cite{henriques2014high}, SRDCF\cite{danelljan2015learning}, Siamese-FC\cite{bertinetto2016fully}, CFNet\cite{valmadre2017end}, MCFTS\cite{liu2017deep}, ECO-deep\cite{danelljan2017eco}, HSSNet\cite{li2019hierarchical}, MLSSNet\cite{liu2020learning}, ECO-MM\cite{liu2022learning}, MMNet\cite{liu2022learning}, ECO-TIR\cite{zhang2018synthetic}, and the latest SiamMSS\cite{li2022multigroup}, DFG\cite{yang2024learning}. 

On the PTB-TIR benchmark, NLMTrack surpasses other trackers in all metrics, achieving a Suc of 0.693 and a Pre of 0.866. Our tracker improves by 5\% in Suc over SiamMSS\cite{li2022multigroup}  and by 1.2\% in Pre over ECO-MM\cite{liu2022learning}, which is the second best on this metric. These results show the robustness and superiority of NLMTrack, which outperforms other trackers significantly. 

\subsubsection{Attribute-Based Evaluation}
\begin{figure*}[!t]
    \centering % 表示居中
    \includegraphics[width=10cm,height=10cm]{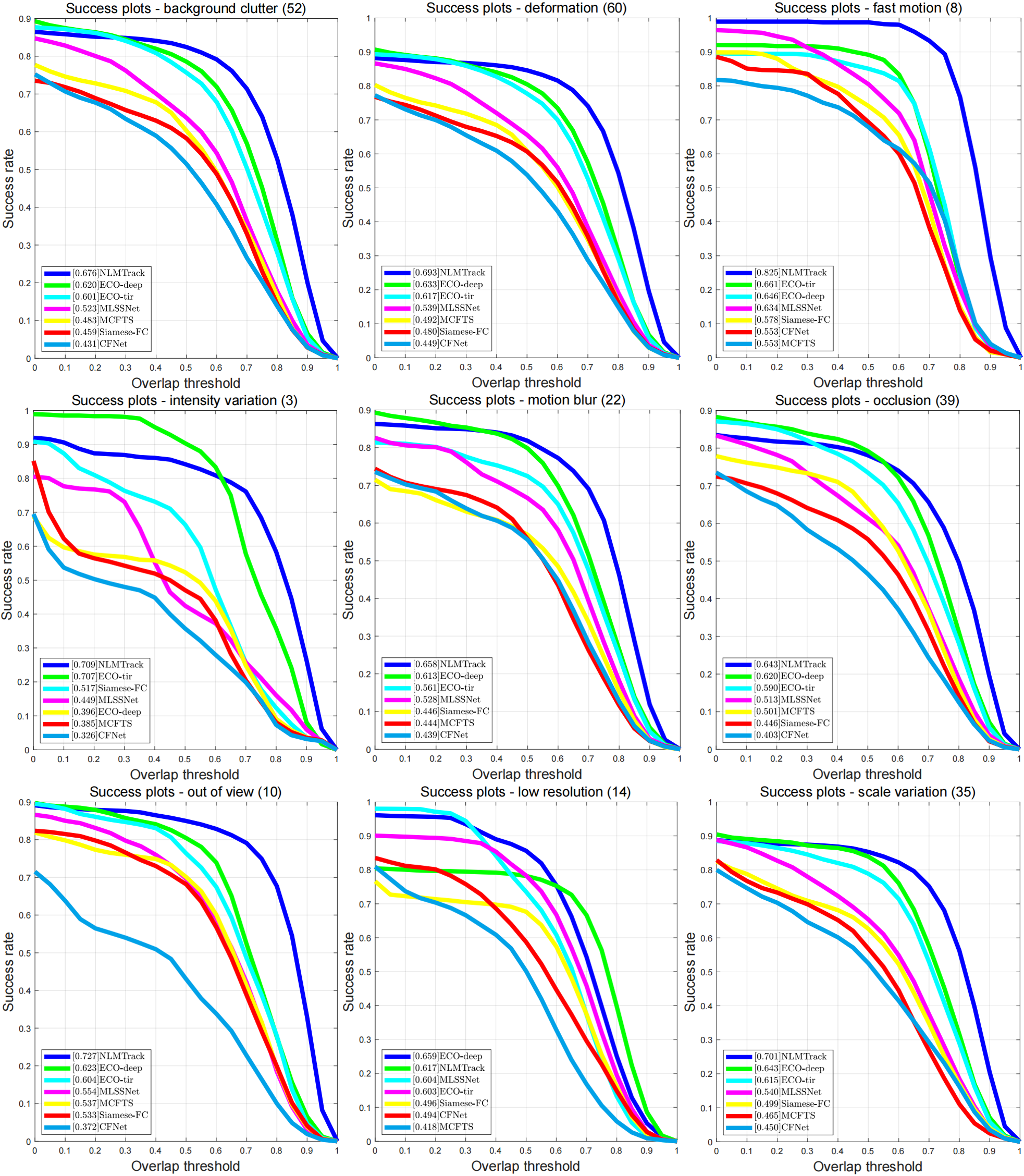}
    \caption{Attribute-based evaluation of NLMTrack on PTB-TIR benchmark.} \label{ptb_attribute}
\end{figure*}

Our NLMTrack tracker is subjected to nine distinct challenge attributes to rigorously evaluate its robustness. Fig. \ref{ptb_attribute}. shows that NLMtrack surpasses other trackers in almost all challenges. NLMTrack achieves a Suc of 5.8\%, 16.4\%, and 5.6\%  higher than the runner-up tracker in scale change, fast motion, and background interference, respectively. This is mainly attributed to the fact that the multilevel progressive fusion module can extract rich semantic information and thus separate the target and the background effectively, and adapt to the scale change of targets with the multiscale features. Moreover, the NLMTrack tracker also surpasses the runner-up tracker by 2.3\% and 6\% in occlusion and deformation, respectively, which indicates that the temporal information in TIR object tracking can overcome the appearance variations.
\begin{table}[!t]
  \centering
  \caption{Evaluation results of NLMtrack and 13 other trackers on PTB-TIR benchmark.}
    \begin{tabular}{l|p{5.625em}ll}
    \toprule[1pt] 
    \multicolumn{1}{p{11.375em}|}{Category} & Tracker & \multicolumn{1}{p{0.5em}}{Suc↑} & \multicolumn{1}{p{0.19em}}{Pre↑} \\
    \midrule
    \multicolumn{1}{l|}{\multirow{2}[2]{*}{Traditional method based trackers}} & KCF\cite{henriques2014high} & 0.400   & 0.595 \\
          & SRDCF\cite{danelljan2015learning} & 0.593 & 0.804 \\
    \midrule
    \multicolumn{1}{l|}{\multirow{8}[2]{*}{Deep learning based CF trackers}} & CFNet\cite{valmadre2017end} & 0.449 & 0.629 \\
          & MCFTS\cite{liu2017deep} & 0.492 & 0.690 \\
          & \multicolumn{1}{p{6.3em}}{ECO-deep\cite{danelljan2017eco}} & 0.633 & 0.838 \\
          & HSSNet\cite{li2019hierarchical} & 0.468 & 0.680 \\
          & MLSSNet\cite{liu2020learning} & 0.516 & 0.731 \\
          & \multicolumn{1}{p{6.6em}}{ECO-MM\cite{liu2022learning}} & 0.659 & 0.854 \\
          & MMNet\cite{liu2022learning} & 0.557 & 0.783 \\
          & \multicolumn{1}{p{6.6em}}{ECO-TIR\cite{zhang2018synthetic}} & 0.617 & 0.830 \\
    \midrule
    \multicolumn{1}{l|}{\multirow{2}[2]{*}{Siamese network-based deep trackers}} & \multicolumn{1}{p{6.8em}}{Siamese-FC\cite{bertinetto2016fully}} & 0.480  & 0.623 \\
          & SiamMSS\cite{li2022multigroup} & 0.643 & 0.823 \\
    \midrule
    \multicolumn{1}{l|}{\multirow{2}[2]{*}{Transformer based trackers}} & DFG\cite{yang2024learning} & 0.634 & 0.803 \\
          & NLMTrack(ours) & \textbf{0.693} & \textbf{0.866} \\
    \bottomrule[1pt] 
    \end{tabular}%
  \label{ptb}%
\end{table}%

\subsection{Evaluation on LSOTB-TIR Dataset}
\subsubsection{Overall Performance}To further demonstrate the robustness of NLMTrack, we also evaluate it with 13 other competing trackers on the LSOTB-TIR benchmark. These trackers include the commonly used KCF\cite{henriques2014high}, SRDCF\cite{danelljan2015learning}, Siamese-FC\cite{bertinetto2016fully}, CFNet\cite{valmadre2017end}, MCFTS\cite{liu2017deep}, ECO-deep\cite{danelljan2017eco}, HSSNet\cite{li2019hierarchical}, MLSSNet\cite{liu2020learning}, ECO-TIR\cite{zhang2018synthetic}, ECO-stir [18], and the newest ones, GFSNet-R\cite{chen2022gfsnet}, SiamMSS\cite{li2022multigroup}, and DFG\cite{yang2024learning}. 

NLMTrack also obtains the best performance in all the metrics, as shown in Table \ref{lsotbtir}, with a Suc of 0.716, a Pre of 0.853, and a NormP of 0.767. Compared to the DFG\cite{yang2024learning} tracker, NLMTrack improves the success rate, the precision rate, and the normalized precision rate by 2.3\%, 2.8\%, and 2.6\%, respectively. Unlike the DFG\cite{yang2024learning} tracker, which only uses a Transformer for feature aggregation, our tracker uses a Transformer to unify feature extraction and feature aggregation, which enhances the discriminative power of the network, and our tracker also introduces temporal and coordinate information as the complementary information of TIR features.
\begin{table}[!t]
  \centering
  \caption{Evaluation results of NLMtrack and 13 other trackers on LSOTB-TIR benchmark. The notation “-” represents no corresponding results given by the authors.}
    \begin{tabular}{l|p{5.625em}lll}
    \toprule[1pt]
    \multicolumn{1}{p{9em}|}{Category} & Tracker & \multicolumn{1}{p{0.5em}}{Suc↑} & \multicolumn{1}{p{0.5em}}{Pre↑} & \multicolumn{1}{p{2.5em}}{NormP↑} \\
    \midrule
    \multicolumn{1}{l|}{\multirow{2}[2]{*}{\begin{tabular}[c]{@{}l@{}}Traditional method \\ based trackers\end{tabular}}} & KCF\cite{henriques2014high} & 0.321 & 0.418 & 0.385 \\
          & SRDCF\cite{danelljan2015learning} & 0.529 & 0.642 & 0.574 \\
    \midrule
    \multicolumn{1}{l|}{\multirow{7}[2]{*}{\begin{tabular}[c]{@{}l@{}}Deep learning based \\ CF trackers\end{tabular}}} & CFNet\cite{valmadre2017end} & 0.416 & 0.519 & 0.481 \\
          & MCFTS\cite{liu2017deep} & 0.479 & 0.635 & 0.546 \\
          & \multicolumn{1}{p{6.5em}}{ECO-deep\cite{danelljan2017eco}} & 0.609 & 0.739 & 0.670 \\
          & \multicolumn{1}{p{6.5em}}{ECO-stir[18]} & 0.616 & 0.750  & 0.672 \\
          & HSSNet\cite{li2019hierarchical} & 0.409 & 0.515 & 0.488 \\
          & MLSSNet\cite{liu2020learning}  & 0.459 & 0.596 & 0.549 \\
          & \multicolumn{1}{p{6.5em}}{ECO-TIR\cite{zhang2018synthetic}} & 0.631 & 0.768 & 0.695 \\
    \midrule
    \multicolumn{1}{l|}{\multirow{3}[2]{*}{\begin{tabular}[c]{@{}l@{}}Siamese network-based \\ deep trackers\end{tabular}}} & \multicolumn{1}{p{6.8em}}{Siamese-FC\cite{bertinetto2016fully}} & 0.517 & 0.651 & 0.587 \\
          & \multicolumn{1}{p{6.625em}}{GFSNet-R\cite{chen2022gfsnet}} & 0.683 & 0.798 & - \\
          & SiamMSS\cite{li2022multigroup} & 0.638 & 0.759 & 0.690 \\
    \midrule
    \multicolumn{1}{l|}{\multirow{2}[2]{*}{Transformer based trackers}} & DFG\cite{yang2024learning} & 0.693 & 0.825 & 0.741 \\
          & NLMTrack(ours) & \textbf{0.716} & \textbf{0.853} & \textbf{0.767} \\
    \bottomrule[1pt]
    \end{tabular}%
  \label{lsotbtir}%
\end{table}%

\subsubsection{Attribute-Based Evaluation}
\begin{figure*}[!t]
    \centering % 表示居中
    \includegraphics[width=12cm,height=10cm]{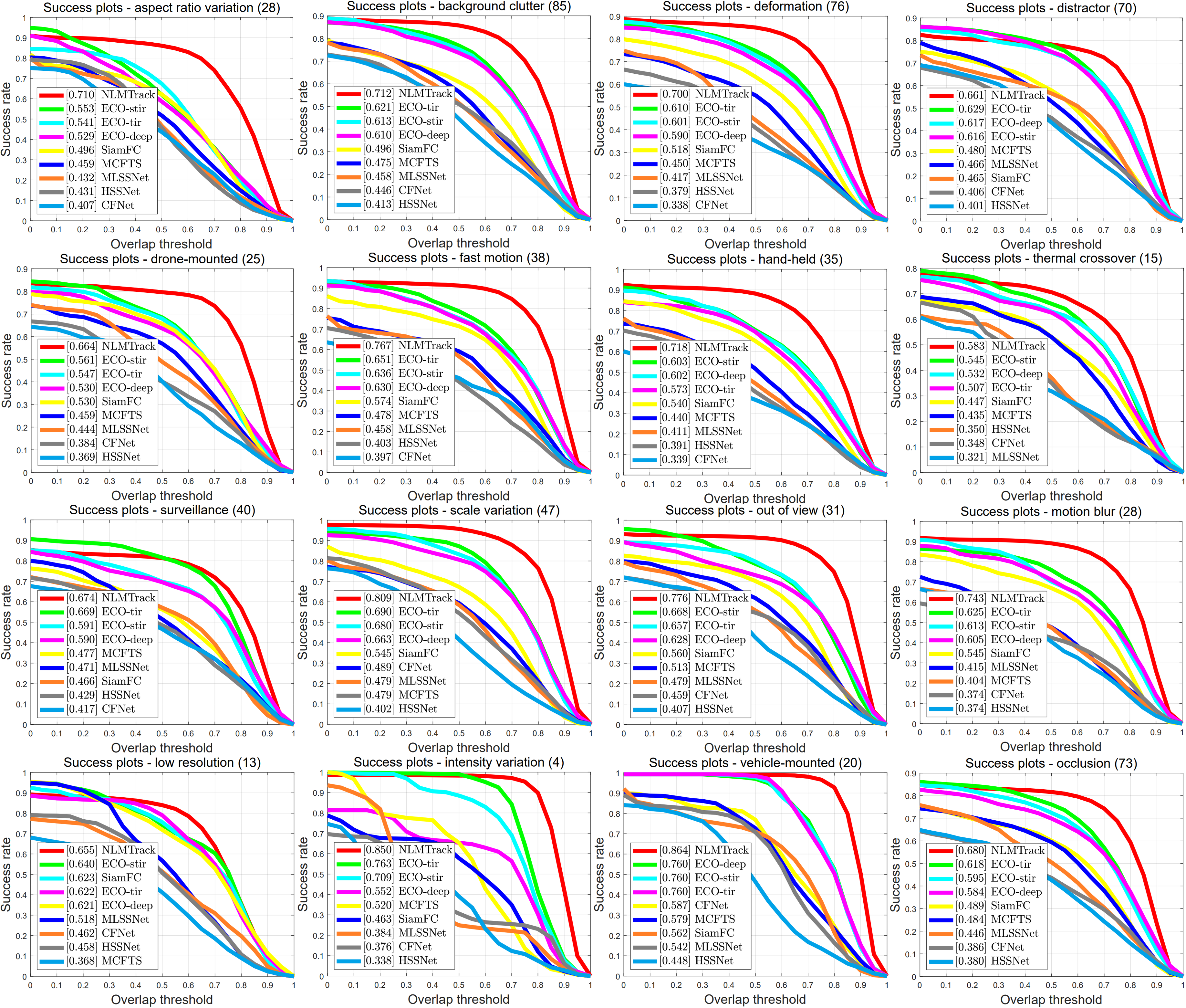}
    \caption{Attribute-based evaluation of NLMTrack on LSOTB-TIR benchmark.} \label{lsotb_attribute}
\end{figure*}

To further analyze the performance of NLMTrack against different challenges, we contrasted NLMTrack with other trackers on 16 different challenge attributes. Compared to the PTB-TIR benchmark, we include seven more challenge attributes, such as thermal crossover, aspect ratio variation, vehicle-mounted, and UAV. Fig. \ref{lsotb_attribute}. shows that NLMTrack outperforms other trackers in all challenges. In the five challenges of scale change, fast motion, background interference, occlusion, and deformation, our tracker improves 11.9\%, 21.6\%, 9.1\%, 6.2\%, and 9\% on success rates over the second-best tracker, respectively, which suggests that NLMTrack achieves a more significant advantage in multiple challenge attributes on a larger dataset, further validating the superiority of NLMTrack based on coordinate sequence generation. Moreover, in the seven additional challenge attributes, such as thermal crossover, aspect ratio variation, vehicle-mounted, UAV, surveillance, handheld, and jamming, NLMTrack surpasses the second-best tracker by 3.8\%, 15.7\%, 10.4\%, 10.3\%, 0.5\%, 11.5\%, and 3.2\% on success rate, respectively. This demonstrates the robustness of NLMTrack based on coordinate sequence generation.

\subsection{Visualized Result}
\begin{figure*}[!t]
\centering
    \subfloat[]{
        \includegraphics[width=2.5cm,height=5cm]{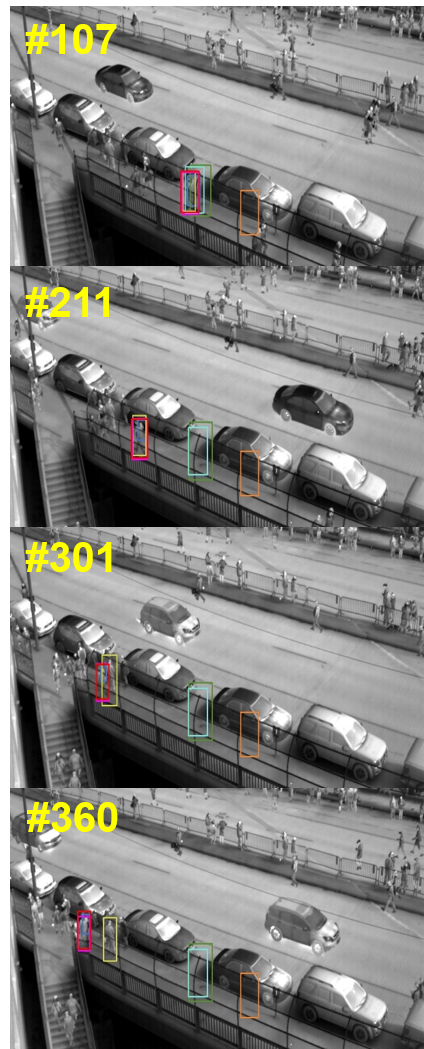} %
        \label{visual(a)}} 
    \subfloat[]{
        \includegraphics[width=2.5cm,height=5cm]{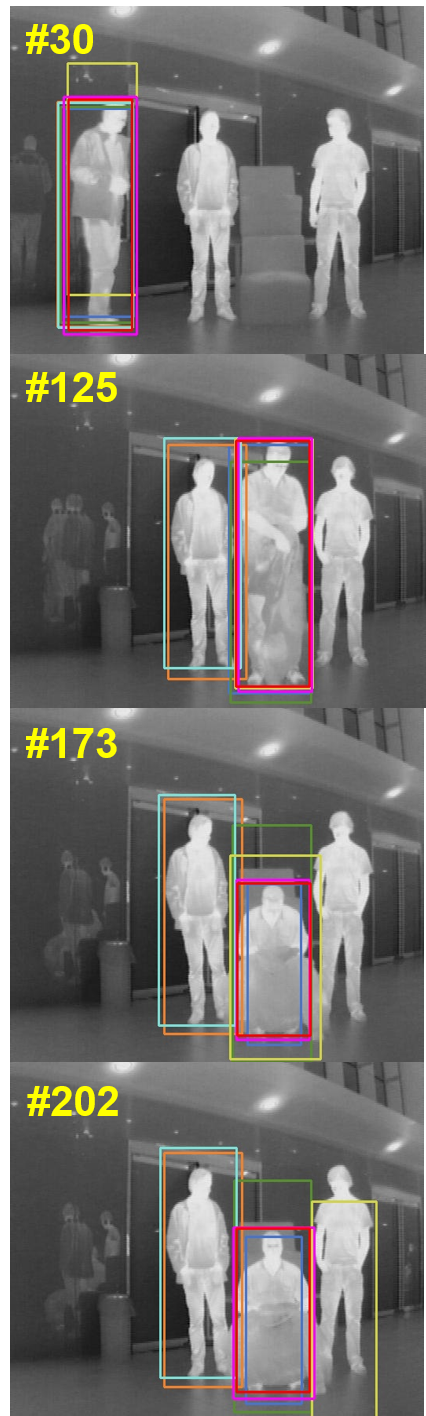} %
        \label{visual(b)}}
    \subfloat[]{
        \includegraphics[width=2.5cm,height=5cm]{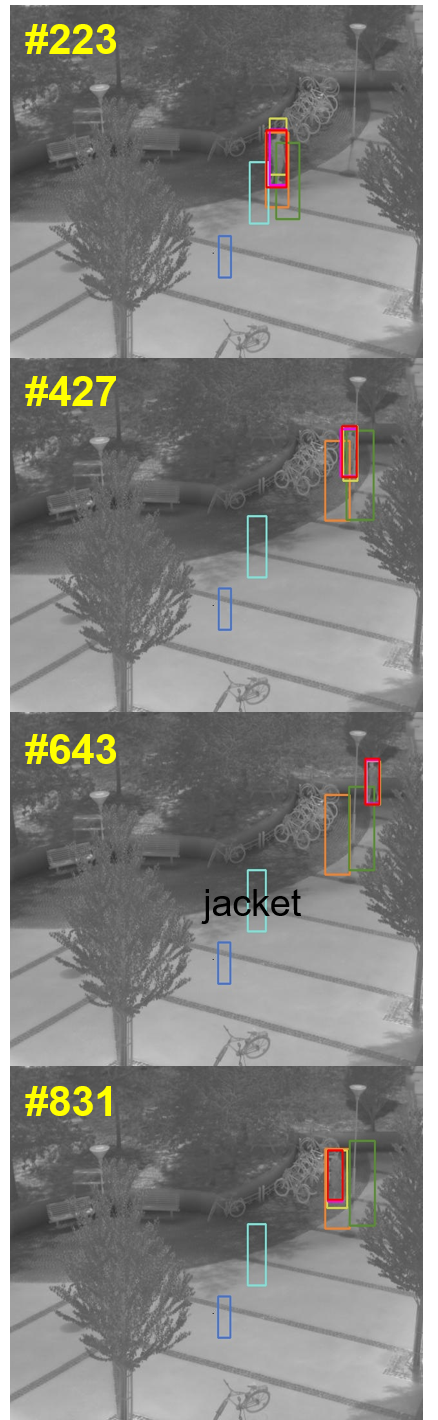} %
        \label{visual(c)}}   
    \subfloat[]{
        \includegraphics[width=2.5cm,height=5cm]{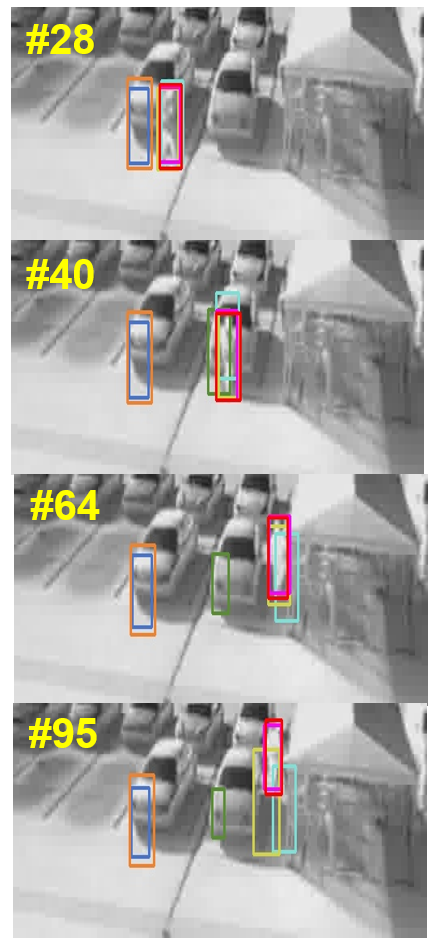} %
        \label{visual(d)}}  
    \subfloat[]{
        \includegraphics[width=2.5cm,height=5cm]{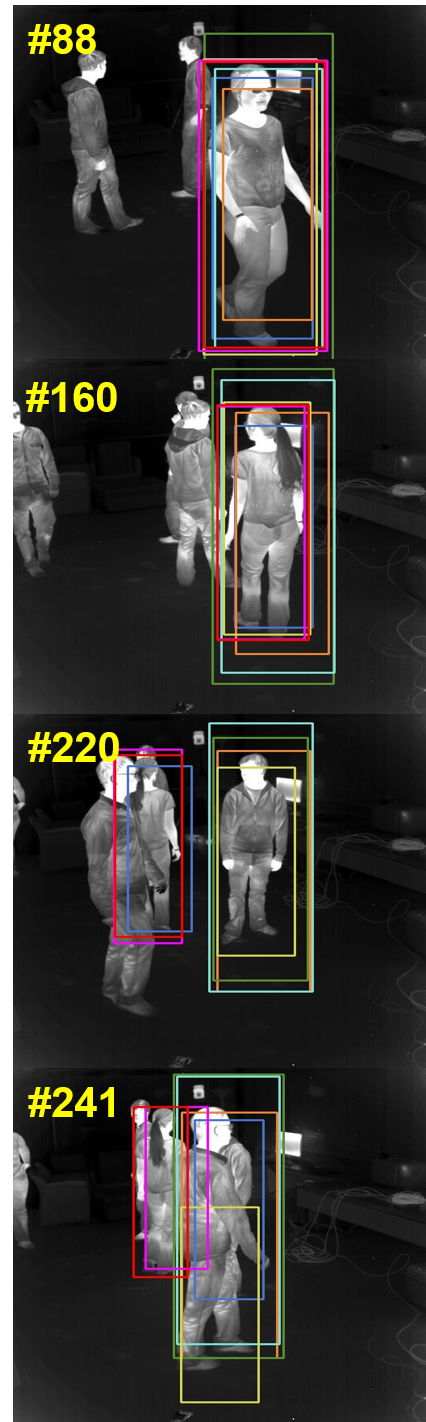} %
        \label{visual(e)}}  
    \subfloat[]{
        \includegraphics[width=2.5cm,height=5cm]{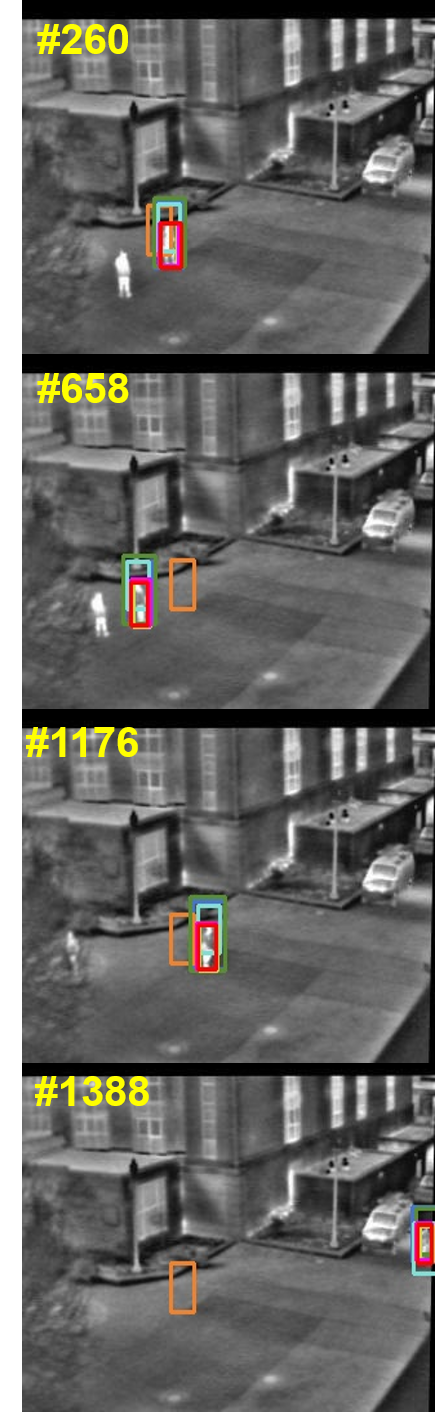} %
        \label{visual(f)}}  
    \subfloat[]{
        \includegraphics[width=2.5cm,height=5cm]{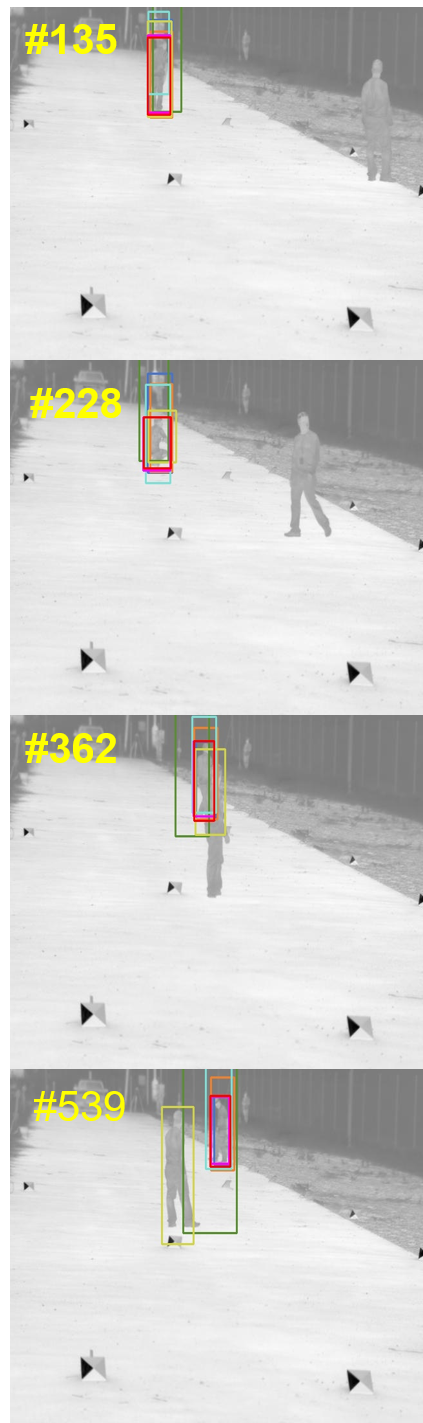} %
        \label{visual(g)}}  
    \quad
    \subfloat{
    \includegraphics[width=10cm,height=0.5cm]{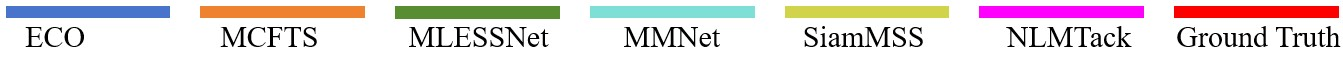} %
    \label{visual(h)}}
    
    \caption{Visualized tracking results of NLMTrack on several challenging sequences. The yellow numbers in the image are the number of frames. } 
    \label{visual}
\end {figure*}

The visualized results of NLMTrack against the five other trackers on the PTB-TIR dataset are demonstrated, as shown in Fig. \ref{visual}. These five trackers are ECO [55], MCFTS\cite{liu2017deep}, MLSSNet\cite{liu2020learning}, MMNet\cite{liu2022learning}, and SiamMSS\cite{li2022multigroup}. Observations from Fig. \ref{visual}\subref{visual(a)}, Fig. \ref{visual}\subref{visual(d)}, and Fig. \ref{visual}\subref{visual(e)} reveal that, in scenarios of occlusion, rival trackers tend to confuse similar objects for the target. In contrast, NLMTrack maintains precise tracking, which is mainly attributed to the fact that NLMTrack can fuse the timing information with the coordinate information to capture changes in the appearance of targets dynamically. Furthermore, as illustrated in Fig. \ref{visual}\subref{visual(b)} and Fig. \ref{visual}\subref{visual(g)}, NLMTrack effectively overcomes interference from similar objects, and it can adapt to change in the target scale promptly compared to the other five trackers. Additionally, Fig. \ref{visual}\subref{visual(c)} demonstrates that NLMTrack can still track the target stably in complex backgrounds, while the other trackers are prone to tracking failures. Finally, as depicted in Fig. \ref{visual}\subref{visual(d)} and Fig.\ref{visual}\subref{visual(f)}, NLMTrack indicates excellent performance in both short-term tracking and long-term tracking.

\subsection{Ablation Study}
The ablation experiments are conducted on the LSOTB-TIR evaluation benchmark. We use Suc, Pre, and NormP as evaluation metrics. 

\subsubsection{Network Architecture}We compare NLMTrack with its three variants in Table \ref{ablation1}: 1)baseline, which is our underlying network architecture that consists of only encoders and decoders. It uses only the cross entropy loss during training and does not use the dynamic template update strategy during testing; 2) baseline+DTUS, which is based on the baseline, and employs a dynamic template during testing to adapt to the target changes; 3) baseline+DTUS+CE+SIOU, which is based on the baseline+DTUS, and uses the loss that combines the cross entropy and SIOU; 4) NLMTrack, which adds a multilevel progressive fusion module (MLPFM) on top of the baseline, and uses the cross-entropy and SIOU combined loss function for supervised learning during training, and the dynamic template update strategy during testing.

A comparative analysis of Table \ref{ablation1} reveals that implementing a dynamic template updating strategy yields a notable increase in tracking performance. Specifically, the Suc metric is enhanced by 1.9\%, while Pre and NormP metrics are enhanced by 2.7\% and 2.4\%, respectively, when comparing the first and second rows. This enhancement underscores the strategy's capacity to effectively harness temporal information, thereby facilitating timely adaptations to the target's appearance changes and ensuring the prompt updating of a reliable target template, which in turn enhances the tracker's stability. Further analysis between the second and third rows in Table \ref{ablation1} suggests that incorporating SIOU loss boosts the Suc by 0.7\% and the Pre and NormP by 0.4\% and 0.3\%, respectively. This proves that the SIOU loss, a priori knowledge that considers the bounding box spatial properties, enhances the performance of the tracker. By comparing the third and last rows in Table \ref{ablation1}, we notice that the Suc is increased by 1.3\%, and the Pre and NormP are increased by 1.1\% and 1.2\%, respectively. This is mainly due to the multilevel progressive fusion module that introduces multiscale features, enriches the semantic information of the network, and thus helps the model to discriminate between target and background more effectively.

\begin{table}[!t]
  \centering
  \caption{Ablation study of NLMTrack on LSOTB-TIR benchmark. DTUS is the dynamic template updating strategy, CE+SIOU refers to the loss function that combines cross-entropy and SIOU, and MLPFM denotes a multilevel progressive fusion module.}
    \begin{tabular}{llllllll}
    \toprule[1pt]
    Tracker & \multicolumn{1}{r}{} & \multicolumn{1}{r}{} & \multicolumn{1}{r}{} &       & \multicolumn{2}{p{5.13em}}{LSOTB-TIR} &  \\
\cmidrule{1-4}\cmidrule{6-8}    Baseline & DTUS  & CE+SIOU & MLPFM &       & \multicolumn{1}{p{0.5em}}{Suc↑} & \multicolumn{1}{p{0.5em}}{Pre↑} & \multicolumn{1}{p{0.5em}}{NormP↑} \\
    \midrule
    \checkmark    & \multicolumn{1}{r}{} & \multicolumn{1}{r}{} & \multicolumn{1}{r}{} &       & 0.667 & 0.800 & 0.718 \\
    \checkmark   & \checkmark    & \multicolumn{1}{r}{} & \multicolumn{1}{r}{} &       & 0.696 & 0.838 & 0.752 \\
    \checkmark    & \checkmark    & \checkmark   & \multicolumn{1}{r}{} &       & 0.703 & 0.842 & 0.755 \\
    \checkmark   & \checkmark    & \checkmark    & \checkmark    &       & \textbf{0.716} & \textbf{0.853} & \textbf{0.767} \\
    \bottomrule[1pt]
    \end{tabular}%
  \label{ablation1}%
\end{table}%

\subsubsection{Fusion Approach}To validate the effectiveness of the progressive fusion approach, we conduct a comparative analysis with two prevalent fusion methodologies. Fig. \ref{fusion}\subref{fusion(a)} shows the first fusion approach (ConF), which fuses multiscale feature maps by a concatenation operation. Fig. \ref{fusion}\subref{fusion(b)} shows the second fusion approach (AddF), which combines multiscale feature maps by an element-wise addition operation. Fig. \ref{fusion}\subref{fusion(c)} shows our fusion approach, which incrementally fuses multilevel feature maps by Upfusion and Downfusion modules. Table \ref{fusiontable} shows that our feature fusion method improves by 0.9\%, 0.7\%, and 0.9\% in Suc, Pre, and NormP, respectively, over the first feature fusion method and improves 0.7\%, 0.6\%, and 0.6\% in Suc, Pre, and NormP, respectively, over the second feature fusion method. This indicates that both the concatenation fusion approach and the element-wise addition fusion approach ignore the semantic gap between different levels and scales, and the direct fusion of these features limits the ability to characterize multiscale features. In contrast, we incrementally fuse semantic features at different levels through the UpFusion module top-down and DownFusion module bottom-up, boosting the tracker's feature representation ability and robustness.

\begin{table}[!t]
  \centering
  \caption{Comparison of different fusion approaches on LSOTB-TIR benchmark.}
    \begin{tabular}{p{4em}llll}
    \toprule[1pt]
    \multicolumn{1}{r}{} & \multicolumn{1}{p{3.75em}}{Suc↑} & \multicolumn{1}{p{4.125em}}{Pre↑} & \multicolumn{1}{p{4.19em}}{NormP↑} \\
    \midrule
    ConF  & 0.707 & 0.846 & 0.758 \\
    AddF  & 0.709 & 0.847 & 0.761 \\
    Ours  & \textbf{0.716} & \textbf{0.853} & \textbf{0.767} \\
    \bottomrule[1pt]
    \end{tabular}%
  \label{fusiontable}%
\end{table}%

\section{Conclusion}
We apply natural language modeling to TIR object tracking for the first time and propose a novel coordinate-aware TIR object tracking model (NLMTrack) in this paper. NLMTrack consists of a Transformer-based encoder, a multilevel progressive fusion module, and a decoder with a causal Transformer. The Transformer-based encoder unifies feature extraction and fusion. The multilevel progressive fusion module enriches the semantic information and introduces multi-scale features, and it is a plug-and-play module. The decoder with a causal Transformer uses the coordinate information and visual features to generate coordinate sequences step-by-step. Furthermore, we devise an adaptive loss function aimed at elevating tracking accuracy, alongside a straightforward template update mechanism that dynamically adjusts to the target's appearance variations. Experimental results on three TIR benchmarks demonstrate that NLMTrack outperforms existing state-of-the-art TIR object trackers. Our NLMTrack provides a promising and effective solution for TIR object tracking, and this “interface” has the potential to be extended to a multi-task learning framework.

%{\appendices
%\section*{Proof of the First Zonklar Equation}
%Appendix one text goes here.
% You can choose not to have a title for an appendix if you want by leaving the argument blank
%\section*{Proof of the Second Zonklar Equation}
%Appendix two text goes here.}

 % argument is your BibTeX string definitions and bibliography database(s)
%\bibliography{IEEEabrv,../bib/paper}

\bibliographystyle{IEEEtran}
\bibliography{main.bib}

% Generated by IEEEtran.bst, version: 1.14 (2015/08/26)
\begin{thebibliography}{10}
\providecommand{\url}[1]{#1}
\csname url@samestyle\endcsname
\providecommand{\newblock}{\relax}
\providecommand{\bibinfo}[2]{#2}
\providecommand{\BIBentrySTDinterwordspacing}{\spaceskip=0pt\relax}
\providecommand{\BIBentryALTinterwordstretchfactor}{4}
\providecommand{\BIBentryALTinterwordspacing}{\spaceskip=\fontdimen2\font plus
\BIBentryALTinterwordstretchfactor\fontdimen3\font minus \fontdimen4\font\relax}
\providecommand{\BIBforeignlanguage}[2]{{%
\expandafter\ifx\csname l@#1\endcsname\relax
\typeout{** WARNING: IEEEtran.bst: No hyphenation pattern has been}%
\typeout{** loaded for the language `#1'. Using the pattern for}%
\typeout{** the default language instead.}%
\else
\language=\csname l@#1\endcsname
\fi
#2}}
\providecommand{\BIBdecl}{\relax}
\BIBdecl

\bibitem{zhou2021object}
Z.~Zhou, X.~Li, T.~Zhang, H.~Wang, and Z.~He, ``Object tracking via spatial-temporal memory network,'' \emph{IEEE Transactions on Circuits and Systems for Video Technology}, vol.~32, no.~5, pp. 2976--2989, 2021.

\bibitem{pi2021instance}
Z.~Pi, Y.~Shao, C.~Gao, and N.~Sang, ``Instance-based feature pyramid for visual object tracking,'' \emph{IEEE Transactions on Circuits and Systems for Video Technology}, vol.~32, no.~6, pp. 3774--3787, 2021.

\bibitem{jian2004real}
C.~Jian and Y.~Jie, ``Real-time infrared object tracking based on mean shift,'' in \emph{Progress in Pattern Recognition, Image Analysis and Applications: 9th Iberoamerican Congress on Pattern Recognition, CIARP 2004, Puebla, Mexico, October 26-29, 2004. Proceedings 9}.\hskip 1em plus 0.5em minus 0.4em\relax Springer, 2004, pp. 45--52.

\bibitem{venkataraman2012adaptive}
V.~Venkataraman, G.~Fan, J.~P. Havlicek, X.~Fan, Y.~Zhai, and M.~B. Yeary, ``Adaptive kalman filtering for histogram-based appearance learning in infrared imagery,'' \emph{IEEE transactions on image processing}, vol.~21, no.~11, pp. 4622--4635, 2012.

\bibitem{liu2017deep}
Q.~Liu, X.~Lu, Z.~He, C.~Zhang, and W.-S. Chen, ``Deep convolutional neural networks for thermal infrared object tracking,'' \emph{Knowledge-Based Systems}, vol. 134, pp. 189--198, 2017.

\bibitem{liu2020learning}
Q.~Liu, X.~Li, Z.~He, N.~Fan, D.~Yuan, and H.~Wang, ``Learning deep multi-level similarity for thermal infrared object tracking,'' \emph{IEEE Transactions on Multimedia}, vol.~23, pp. 2114--2126, 2020.

\bibitem{li2022multigroup}
W.~Li, L.~Lv, and J.~Zhu, ``Multigroup spatial shift models for thermal infrared tracking,'' \emph{Knowledge-Based Systems}, vol. 255, p. 109705, 2022.

\bibitem{yang2024learning}
C.~Yang, Q.~Liu, G.~Li, H.~Pan, and Z.~He, ``Learning diverse fine-grained features for thermal infrared tracking,'' \emph{Expert Systems with Applications}, vol. 238, p. 121577, 2024.

\bibitem{chen2021pix2seq}
T.~Chen, S.~Saxena, L.~Li, D.~J. Fleet, and G.~Hinton, ``Pix2seq: A language modeling framework for object detection,'' \emph{arXiv preprint arXiv:2109.10852}, 2021.

\bibitem{chen2023seqtrack}
X.~Chen, H.~Peng, D.~Wang, H.~Lu, and H.~Hu, ``Seqtrack: Sequence to sequence learning for visual object tracking,'' in \emph{Proceedings of the IEEE/CVF Conference on Computer Vision and Pattern Recognition}, 2023, pp. 14\,572--14\,581.

\bibitem{liu2019ptb}
Q.~Liu, Z.~He, X.~Li, and Y.~Zheng, ``Ptb-tir: A thermal infrared pedestrian tracking benchmark,'' \emph{IEEE Transactions on Multimedia}, vol.~22, no.~3, pp. 666--675, 2019.

\bibitem{felsberg2015thermal}
M.~Felsberg, A.~Berg, G.~Hager, J.~Ahlberg, M.~Kristan, J.~Matas, A.~Leonardis, L.~Cehovin, G.~Fernandez, T.~Vojir \emph{et~al.}, ``The thermal infrared visual object tracking vot-tir2015 challenge results,'' in \emph{Proceedings of the ieee international conference on computer vision workshops}, 2015, pp. 76--88.

\bibitem{liu2023lsotb}
Q.~Liu, X.~Li, D.~Yuan, C.~Yang, X.~Chang, and Z.~He, ``Lsotb-tir: A large-scale high-diversity thermal infrared single object tracking benchmark,'' \emph{IEEE Transactions on Neural Networks and Learning Systems}, 2023.

\bibitem{gao2016infrared}
S.~J. Gao and S.~T. Jhang, ``Infrared target tracking using multi-feature joint sparse representation,'' in \emph{Proceedings of the International Conference on Research in Adaptive and Convergent Systems}, 2016, pp. 40--45.

\bibitem{berg2016channel}
A.~Berg, J.~Ahlberg, and M.~Felsberg, ``Channel coded distribution field tracking for thermal infrared imagery,'' in \emph{Proceedings of the IEEE Conference on computer vision and pattern recognition workshops}, 2016, pp. 9--17.

\bibitem{ding2019thermal}
M.~Ding, X.~Zhang, W.-H. Chen, L.~Wei, and Y.-F. Cao, ``Thermal infrared pedestrian tracking via fusion of features in driving assistance system of intelligent vehicles,'' \emph{Proceedings of the Institution of Mechanical Engineers, Part G: Journal of Aerospace Engineering}, vol. 233, no.~16, pp. 6089--6103, 2019.

\bibitem{yun2019tir}
S.~Yun and S.~Kim, ``Tir-ms: Thermal infrared mean-shift for robust pedestrian head tracking in dynamic target and background variations,'' \emph{Applied Sciences}, vol.~9, no.~15, p. 3015, 2019.

\bibitem{demir2016co}
H.~S. Demir and A.~E. Cetin, ``Co-difference based object tracking algorithm for infrared videos,'' in \emph{2016 IEEE International Conference on Image Processing (ICIP)}.\hskip 1em plus 0.5em minus 0.4em\relax IEEE, 2016, pp. 434--438.

\bibitem{liu2022learning}
Q.~Liu, D.~Yuan, N.~Fan, P.~Gao, X.~Li, and Z.~He, ``Learning dual-level deep representation for thermal infrared tracking,'' \emph{IEEE Transactions on Multimedia}, vol.~25, pp. 1269--1281, 2022.

\bibitem{zhang2018synthetic}
L.~Zhang, A.~Gonzalez-Garcia, J.~Van De~Weijer, M.~Danelljan, and F.~S. Khan, ``Synthetic data generation for end-to-end thermal infrared tracking,'' \emph{IEEE Transactions on Image Processing}, vol.~28, no.~4, pp. 1837--1850, 2018.

\bibitem{danelljan2017eco}
M.~Danelljan, G.~Bhat, F.~Shahbaz~Khan, and M.~Felsberg, ``Eco: Efficient convolution operators for tracking,'' in \emph{Proceedings of the IEEE conference on computer vision and pattern recognition}, 2017, pp. 6638--6646.

\bibitem{li2019hierarchical}
X.~Li, Q.~Liu, N.~Fan, Z.~He, and H.~Wang, ``Hierarchical spatial-aware siamese network for thermal infrared object tracking,'' \emph{Knowledge-Based Systems}, vol. 166, pp. 71--81, 2019.

\bibitem{ding2022thermal}
M.~Ding, W.-H. Chen, and Y.-F. Cao, ``Thermal infrared single-pedestrian tracking for advanced driver assistance system,'' \emph{IEEE Transactions on Intelligent Vehicles}, vol.~8, no.~1, pp. 814--824, 2022.

\bibitem{parhizkar2023object}
M.~Parhizkar, G.~Karamali, and B.~Abedi~Ravan, ``Object tracking in infrared images using a deep learning model and a target-attention mechanism,'' \emph{Complex \& Intelligent Systems}, vol.~9, no.~2, pp. 1495--1506, 2023.

\bibitem{yuan2023robust}
D.~Yuan, X.~Shu, Q.~Liu, X.~Zhang, and Z.~He, ``Robust thermal infrared tracking via an adaptively multi-feature fusion model,'' \emph{Neural Computing and Applications}, vol.~35, no.~4, pp. 3423--3434, 2023.

\bibitem{bertinetto2016fully}
L.~Bertinetto, J.~Valmadre, J.~F. Henriques, A.~Vedaldi, and P.~H. Torr, ``Fully-convolutional siamese networks for object tracking,'' in \emph{Computer Vision--ECCV 2016 Workshops: Amsterdam, The Netherlands, October 8-10 and 15-16, 2016, Proceedings, Part II 14}.\hskip 1em plus 0.5em minus 0.4em\relax Springer, 2016, pp. 850--865.

\bibitem{chen2022gfsnet}
R.~Chen, S.~Liu, Z.~Miao, and F.~Li, ``Gfsnet: Generalization-friendly siamese network for thermal infrared object tracking,'' \emph{Infrared Physics \& Technology}, vol. 123, p. 104190, 2022.

\bibitem{zhao2022thermal}
L.~Zhao, X.~Liu, H.~Ren, and L.~Xue, ``Thermal infrared tracking method based on efficient global information perception,'' \emph{Sensors}, vol.~22, no.~19, p. 7408, 2022.

\bibitem{brown2020language}
T.~Brown, B.~Mann, N.~Ryder, M.~Subbiah, J.~D. Kaplan, P.~Dhariwal, A.~Neelakantan, P.~Shyam, G.~Sastry, A.~Askell \emph{et~al.}, ``Language models are few-shot learners,'' \emph{Advances in neural information processing systems}, vol.~33, pp. 1877--1901, 2020.

\bibitem{sun2019videobert}
C.~Sun, A.~Myers, C.~Vondrick, K.~Murphy, and C.~Schmid, ``Videobert: A joint model for video and language representation learning,'' in \emph{Proceedings of the IEEE/CVF international conference on computer vision}, 2019, pp. 7464--7473.

\bibitem{ding2022cogview2}
M.~Ding, W.~Zheng, W.~Hong, and J.~Tang, ``Cogview2: Faster and better text-to-image generation via hierarchical transformers,'' \emph{Advances in Neural Information Processing Systems}, vol.~35, pp. 16\,890--16\,902, 2022.

\bibitem{gevorgyan2022siou}
Z.~Gevorgyan, ``Siou loss: More powerful learning for bounding box regression,'' \emph{arXiv preprint arXiv:2205.12740}, 2022.

\bibitem{yuan2021temporal}
Z.~Yuan, X.~Song, L.~Bai, Z.~Wang, and W.~Ouyang, ``Temporal-channel transformer for 3d lidar-based video object detection for autonomous driving,'' \emph{IEEE Transactions on Circuits and Systems for Video Technology}, vol.~32, no.~4, pp. 2068--2078, 2021.

\bibitem{wei2023autoregressive}
X.~Wei, Y.~Bai, Y.~Zheng, D.~Shi, and Y.~Gong, ``Autoregressive visual tracking,'' in \emph{Proceedings of the IEEE/CVF Conference on Computer Vision and Pattern Recognition}, 2023, pp. 9697--9706.

\bibitem{cui2022mixformer}
Y.~Cui, C.~Jiang, L.~Wang, and G.~Wu, ``Mixformer: End-to-end tracking with iterative mixed attention,'' in \emph{Proceedings of the IEEE/CVF Conference on Computer Vision and Pattern Recognition}, 2022, pp. 13\,608--13\,618.

\bibitem{li2022exploring}
Y.~Li, H.~Mao, R.~Girshick, and K.~He, ``Exploring plain vision transformer backbones for object detection,'' in \emph{European Conference on Computer Vision}.\hskip 1em plus 0.5em minus 0.4em\relax Springer, 2022, pp. 280--296.

\bibitem{yan2021learning}
B.~Yan, H.~Peng, J.~Fu, D.~Wang, and H.~Lu, ``Learning spatio-temporal transformer for visual tracking,'' in \emph{Proceedings of the IEEE/CVF international conference on computer vision}, 2021, pp. 10\,448--10\,457.

\bibitem{huang2019got}
L.~Huang, X.~Zhao, and K.~Huang, ``Got-10k: A large high-diversity benchmark for generic object tracking in the wild,'' \emph{IEEE transactions on pattern analysis and machine intelligence}, vol.~43, no.~5, pp. 1562--1577, 2019.

\bibitem{gundogdu2015comparison}
E.~Gundogdu, H.~Ozkan, H.~Seckin~Demir, H.~Ergezer, E.~Akagunduz, and S.~Kubilay~Pakin, ``Comparison of infrared and visible imagery for object tracking: Toward trackers with superior ir performance,'' in \emph{Proceedings of the IEEE Conference on Computer Vision and Pattern Recognition Workshops}, 2015, pp. 1--9.

\bibitem{danelljan2015learning}
M.~Danelljan, G.~Hager, F.~Shahbaz~Khan, and M.~Felsberg, ``Learning spatially regularized correlation filters for visual tracking,'' in \emph{Proceedings of the IEEE international conference on computer vision}, 2015, pp. 4310--4318.

\bibitem{valmadre2017end}
J.~Valmadre, L.~Bertinetto, J.~Henriques, A.~Vedaldi, and P.~H. Torr, ``End-to-end representation learning for correlation filter based tracking,'' in \emph{Proceedings of the IEEE conference on computer vision and pattern recognition}, 2017, pp. 2805--2813.

\bibitem{henriques2014high}
J.~F. Henriques, R.~Caseiro, P.~Martins, and J.~Batista, ``High-speed tracking with kernelized correlation filters,'' \emph{IEEE transactions on pattern analysis and machine intelligence}, vol.~37, no.~3, pp. 583--596, 2014.

\end{thebibliography}

\vfill

\end{document}